\documentclass[lettersize,journal]{IEEEtran}
\usepackage{amsmath,amsfonts}
\usepackage{amssymb}
\usepackage{algorithm}
\usepackage{algorithmic}
\usepackage{array}
\usepackage[caption=false,font=footnotesize]{subfig}
\usepackage{textcomp}
\usepackage{stfloats}
\usepackage{multirow}
\usepackage{url}
\usepackage{verbatim}
\usepackage{graphicx}
\usepackage[table]{xcolor} 
\usepackage{cite}
\usepackage[colorlinks=true, linkcolor=blue, citecolor=red, urlcolor=magenta]{hyperref}
\def\BibTeX{{\rm B\kern-.05em{\sc i\kern-.025em b}\kern-.08em
    T\kern-.1667em\lower.7ex\hbox{E}\kern-.125emX}}
\usepackage{balance}
\begin{document}
\title{TRAIN: Terrain-Aware Interactive Navigation for Quadruped Robots via Reinforcement Learning}
\author{Wei Zhu$^1$, Irfan Tito Kurniawan$^1$, Ye Zhao$^2$, and Mitsuhiro Hayashibe$^1$
\thanks{This work was supported in part by the JST Moonshot Research and Development under Grant JPMJMS263H. (\textit{Corresponding author: Mitsuhiro Hayashibe)}}
\thanks{$^1$The authors are with the Department of Robotics, Graduate School of Engineering, Tohoku University, 980-8579, Sendai, Japan. {\tt\small zhu.wei.c1@tohoku.ac.jp}}
\thanks{$^2$The author is with the Laboratory for Intelligent Decision and Autonomous Robots, Woodruff School of Mechanical Engineering, Georgia Institute of Technology, Atlanta, GA 30313, USA. {\tt\small yezhao@gatech.edu}}
\thanks{Project website: https://tohoku-neurobotics.github.io/trance.github.io/.}}

\maketitle

\begin{abstract}
This study introduces TRAIN: TeRrain-Aware Interactive Navigation, a deep reinforcement learning (DRL)-based framework for quadrupedal navigation across unstructured terrain and human-populated environments. Conventional quadrupedal navigation typically separates motion planning from locomotion control, neglecting whole-body constraints and terrain awareness. On the other hand, end-to-end methods are more integrated but require high-frequency sensing, which is often noisy and computationally costly. In addition, most existing approaches assume static environments, limiting their use in human-populated settings. To address these limitations, we propose a two-stage training framework with three DRL pipelines. (1) TRAIN-Loco employs an asymmetric actor–critic (AC) model for quadrupedal locomotion, enabling agile traversal of uneven terrains without explicit terrain or contact observations. (2) TRAIN-Nav applies a symmetric AC framework for navigation with interactive pedestrians, directly mapping transformed LiDAR data to ego-agent actions under differential-drive kinematics. (3) A unified pipeline, TRAIN, integrates TRAIN-Loc and TRAIN-Nav, supporting terrain-aware quadrupedal navigation in uneven and interactive environments. Comprehensive benchmarks against quadrupedal locomotion and robot navigation baselines demonstrate the effectiveness of TRAIN. Hardware experiments further confirm its potential for sim-to-real transfer.

\textit{Note to Practitioners}—This paper is motivated by the challenge of deploying quadrupedal robots in real-world environments that are both physically complex (uneven terrain) and interactively dynamic (populated by humans). While quadrupedal robots excel at traversing difficult landscapes, existing navigation frameworks often decouple high-level planning from low-level gait control, leading to failures when the robot’s physical constraints are ignored. Furthermore, traditional navigation models often assume static human behavior or rely on computationally expensive sensing that is difficult to deploy on physical hardware. This paper introduces TRAIN (TeRrain-Aware Interactive Navigation), a deep reinforcement learning framework that unifies terrain-aware locomotion with interaction-aware navigation for quadruped robots. By utilizing an asymmetric actor–critic model, the robot learns to traverse irregular surfaces without needing precise terrain maps. Simultaneously, a navigation policy with interaction awareness is developed using transformed LiDAR data to capture human-robot interactions implicitly, ensuring the robot moves in a interactively acceptable manner. A key advantage of the TRAIN framework is its hierarchical design, which allows for robust sim-to-real transfer and real-time execution on resource-constrained hardware. Experimental results demonstrate that the system can successfully manage complex interactions while maintaining stable movement over unstructured ground. Practically, this framework provides a scalable solution for using quadrupedal robots in applications such as search and rescue, last-mile delivery, and facility inspection, where robots must seamlessly navigate around people across diverse indoor and outdoor surfaces. Future research will explore the integration of active vision to further improve the robot’s intent recognition in even more crowded or unpredictable interactive settings.
\end{abstract}


\section{Introduction}

\begin{figure}[!t]
	\centering
	\small
	\includegraphics[width=8.5cm]{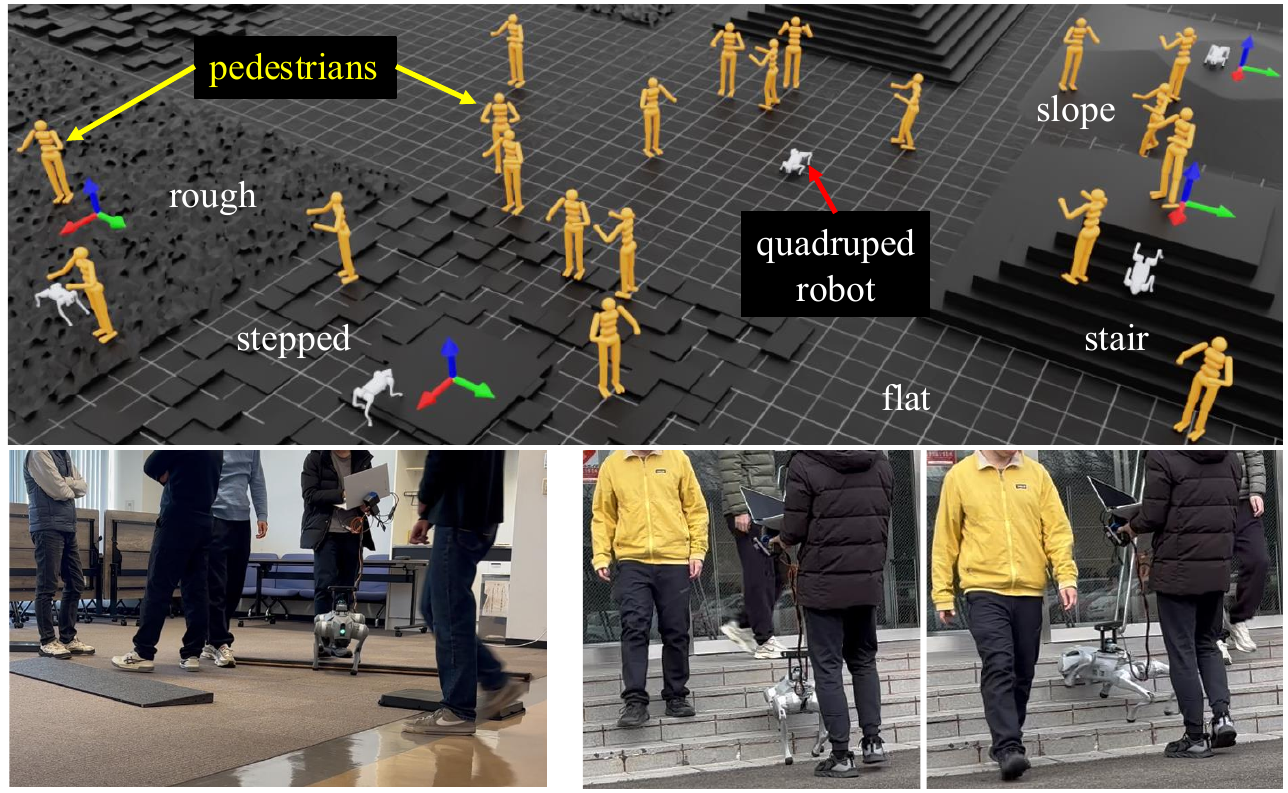}
	\caption{Terrain-aware navigation of quadruped robots in interactive environments. The top panel illustrates simulation training scenarios, while the bottom panels show real-world implementations.}
	\label{introductionFigure}
\end{figure}

\IEEEPARstart{Q}{uadruped} robots exhibit superior maneuverability compared to their wheeled counterparts, allowing them to navigate uneven terrain and execute dynamic maneuvers in complex environments. While locomotion on irregular surfaces has been extensively studied \cite{ha2024learning}, navigation challenges, such as collision avoidance and target reaching in uneven and dynamic environments, remain comparatively underexplored (see Fig.~\ref{introductionFigure}). Such capabilities are essential for deploying agile robots in interactive and real-world settings. Existing navigation frameworks \cite{xie2024real} often decouple motion planning from locomotion control, which tends to neglect critical whole-body constraints. Conversely, while end-to-end frameworks \cite{zhang2024resilient} attempt to unify these layers, they generally demand high-frequency exteroceptive sensing that is often precluded by onboard computational constraints and inherent sensor noise.

Compared to quadrupedal platforms, the navigation of wheeled mobile robots in static environments is well-established \cite{francis2020long, devo2020towards, duong2022autonomous, xiao2023barriernet}, encompassing both model-based planners and model-free pipelines. While the non-evolving nature of static scenes facilitates real-world deployment, navigation in dynamic environments remains a formidable challenge due to complex, implicit agent interactions, such as human–human and human–robot coordination. Current model-based approaches generally assume constant pedestrian velocities and overlook these latent interactions \cite{singamaneni2024survey}. Furthermore, while predictive models can forecast pedestrian trajectories by taking interactions into consideration \cite{alahi2016social}, their practical utility is often hampered by out-of-distribution (OOD) data and robot–model mismatch. Although model-free pipelines offer a potential solution \cite{zhu2021deep}, they typically rely on high-fidelity expert demonstrations or assume omnidirectional kinematics that neglect physical constraints. Moreover, methods reliant on raw sensor observations \cite{fan2020distributed} often face significant sim-to-real gaps due to geometric and perceptual discrepancies between simulated and real-world entities.

In this work, we present \textbf{TRAIN}: \textbf{T}e\textbf{R}rain-\textbf{A}ware \textbf{I}nteractive \textbf{N}avigation for quadruped robots via reinforcement learning. Our approach decouples quadrupedal locomotion from navigation by independently training two distinct DRL policies. We first utilize an asymmetric actor–critic (AC) architecture that operates without explicit terrain maps or contact states. Concurrently, a symmetric AC framework is developed for navigation, employing a differential-drive model governed by kinematic constraints. Subsequently, the navigation policy is fine-tuned and extended for the quadrupedal platform, yielding a unified, hierarchical policy capable of robust quadrupedal navigation in uneven and human-populated environments. 

The primary contributions of this study are as follows:

\begin{itemize}

\item \textbf{Terrain-aware quadrupedal locomotion}. We develop an agile quadrupedal locomotion controller that operates without explicit terrain maps or contact state estimation through an asymmetric AC architecture, thereby streamlining hardware deployment. 

\item \textbf{Navigation in interactive environments with transformed LiDAR scans}. We present a navigation policy that implicitly captures interactive features from transformed LiDAR scans. This approach streamlines the sim-to-real transfer without relying on fully known environments, significantly facilitating real-world implementations.

\item \textbf{Unified navigation and locomotion}. We propose an integrated architecture that harmonizes high-level motion planning with low-level locomotion. This unification allows for the optimization of navigation paths while maintaining strict adherence to the robot's whole-body constraints and terrain awareness.

\item \textbf{Comprehensive benchmarks and hardware evaluations}. We reproduce and extend state-of-the-art (SOTA) benchmarks for quadrupedal locomotion and conduct rigorous comparative analyses against both traditional model-based planners and DRL-based baselines. Finally, we demonstrate the practical efficacy of our framework through successful sim-to-real deployment on a physical quadruped platform.  
    
\end{itemize}

The remainder of this article is organized as follows. Section~\ref{related_work} provides a review of relevant literature. Section~\ref{overall_framework} introduces the proposed framework and formalizes the problem. Section~\ref{quadruped_locomotion} details the locomotion controller, and Section~\ref{social_navigation} presents the navigation policy with interaction awareness. In Section~\ref{unified_quadruped_navigation}, we describe the integrated algorithm for quadrupedal navigation in interactive, uneven environments.
The performance of the locomotion, navigation, and unified frameworks is evaluated through simulation in Sections~\ref{simulation_evaluations_locomotion}, \ref{simulation_evaluations_navigation}, and \ref{simulation_evaluations_unify}, respectively. Section~\ref{hardware_implementations} discusses the technical details of the sim-to-real transfer and hardware deployment. Finally, Section~\ref{discussions} addresses the limitations of the current study, and Section~\ref{conclusions} concludes the paper.

\section{Related Work} \label{related_work}
This section provides a comprehensive review of the relevant literature. We first review navigation strategies for wheeled mobile robots, given their widespread deployment in interactive environments. These strategies are categorized into two primary paradigms: classical model-based approaches and learning-based pipelines. Subsequently, we summarize the state of the research regarding quadrupedal locomotion, focusing specifically on the integration of these controllers into navigation frameworks.

\subsection{Model-based Navigation Approaches}
Model-based navigation for mobile robots in dynamic environments relies on explicit environmental representations, robot dynamics, and predicted agent trajectories to generate control actions. For instance, given the robot’s kinematics (e.g., differential-drive models) and the state information of surrounding agents, such as geometry, position, and velocity, the Dynamic Window Approach (DWA) \cite{fox2002dynamic, dobrevski2024dynamic} executes a discrete search over the feasible action space. However, DWA typically assumes constant pedestrian velocities within a fixed planning horizon, which often results in unnatural maneuvers or collision failures when encountering interactive, time-varying pedestrian behaviors.

Assuming fully observable agent states, the Social Force Model (SFM) \cite{helbing1995social, ferrer2014proactive} and Optimal Reciprocal Collision Avoidance (ORCA) \cite{van2011reciprocal, alonso2013optimal} optimize the trajectories of interacting agents through local force or velocity-space constraints. However, SFM lacks explicit collision avoidance guarantees, while ORCA assumes reciprocity—a condition where all agents share equal responsibility for collision avoidance. Although variants such as Variable Responsibility ORCA (VR-ORCA) \cite{guo2021vr} and AdaptiVe Optimal Collision Avoidance Driven by Opinion (AVOCADO) \cite{martinez2025avocado} relax this assumption, these methods remain inherently reactive. By prioritizing immediate state-based actions over long-horizon trajectory planning, they are prone to oscillations and unnatural detours in dense or symmetric configurations. Additionally, their heavy reliance on precise agent geometry and kinematic data necessitates highly accurate sensing, which is often difficult to maintain in real-world scenarios.

To facilitate long-horizon planning and mitigate unnatural behaviors in symmetric interactions, Model Predictive Control (MPC)-based frameworks have gained significant traction for navigation in dynamic environments. For example, Safe Horizon MPC (SH-MPC) \cite{de2023scenario} explicitly constrains collision probabilities throughout the planned trajectory, while Topology-driven MPC (T-MPC) \cite{de2024topology} integrates global guidance as a topological constraint for local motion planning. Despite these advancements, both approaches often assume constant pedestrian velocities over the planning horizon, which limits their adaptability to complex interactions. 

To address this limitation, the Neural Proximal Alternating-minimization Network (NeuPAN) \cite{han2025neupan} utilizes a Deep Unfolded Neural Encoder (DUNE) to predict collision points directly from raw LiDAR point clouds, which are then used to inform the MPC planner. Beyond raw point cloud data, the Stochastic Cartographic Occupancy Prediction Engine (SCOPE) \cite{xie2025scope} utilizes a Variational Autoencoder (VAE) to predict potential collision regions from occupancy grid maps (OGMs). Similarly, spatiotemporal OGMs have been employed for dynamic scene prediction via self-supervised learning frameworks \cite{thomas2023foreseeable}. However, such predictive models are trained on specific datasets, potentially compromising their generalization to novel environments or unseen pedestrian behaviors. Moreover, MPC-based navigation often exhibits over-conservative behavior, particularly when navigating interactive environments where feasible paths are narrow.

\subsection{Model-free Navigation Pipelines}
In contrast to model-based approaches, model-free navigation pipelines, typically relying on machine learning techniques, offer a more streamlined architecture. This is achieved by bypassing the explicit integration of perception and planning, eliminating the need for online optimization, and reducing the burden of manual parameter tuning. Collision Avoidance with Deep Reinforcement Learning (CADRL) \cite{chen2017decentralized} pioneered this paradigm by directly mapping agent observations to control actions. However, CADRL primarily accounts for pairwise interactions between the robot and individual agents, neglecting the broader collective dynamics of the interactive scenarios. Subsequent frameworks, such as Self-Attention RL (SARL) \cite{chen2019crowd} and Relational Graph Learning (RGL) \cite{chen2020relational}, address this by leveraging attention mechanisms and Graph Neural Networks (GNNs) to model implicit multi-agent interactions. Despite these advancements, such methods often assume perfect state estimation, including agent geometry and preferred velocities, which exacerbates the sim-to-real gap due to their heavy reliance on high-fidelity perception. Moreover, these pipelines frequently require expert demonstrations for pre-training and typically assume omnidirectional kinematics. This lack of consideration for physical motion constraints limits their applicability to complex robotic platforms with non-holonomic or strict kinematic limitations.

An alternative, more direct paradigm utilizes DRL to map raw sensor observations into control actions. For instance, Hybrid-RL \cite{fan2020distributed} incorporates an RL policy alongside two specialized safety and efficiency modules tailored for hazardous and low-complexity scenarios, respectively. In standard environments, the RL agent directly transforms continuous LiDAR scans into velocity commands for a differential-drive robot. While this hierarchical approach offers a balance of safety and performance, the robust classification of scenarios remains a significant challenge. Consequently, recent research has shifted toward developing unified DRL policies capable of generalizing across diverse environments. One map-free navigation policy \cite{jin2020mapless} extracts latent interaction features from sequential LiDAR data, while Learning to Navigate in Dynamic environments with Normalized LiDAR scans (LNDNL) \cite{zhu2024learn} leverages Long Short-Term Memory (LSTM) networks to capture spatiotemporal dynamics. Similarly, Distilling Privileged information for Crowd-Aware Navigation (DiPCAN) \cite{monaci2022dipcan} maps depth imagery to robot actions by distilling privileged pedestrian state information. Beyond raw scans, intermediate representations such as occupancy maps \cite{zhu2023autonomous} and risk maps \cite{yang2023rmrl} are frequently employed to encode interaction geometry. However, high-dimensional image processing entails significant computational overhead, and map-based abstractions often suffer from degraded spatial resolution in expansive environments. Furthermore, the use of continuous LiDAR scans inherently couples ego-motion with pedestrian dynamics, potentially hindering learning efficiency and the asymptotic performance of the policy.

Integrating raw sensor data with privileged pedestrian states, such as precise positions and velocities, enables the extraction of richer interaction features \cite{sathyamoorthy2020densecavoid, yao2021crowd, xie2023drl, Liu2026Height}. However, the practical utility of these methods is often limited by the need for sophisticated perception and estimation pipelines to provide such ground-truth information in real-time. In addition to DRL, Imitation Learning (IL) \cite{qin2021deep, karnan2022socially, nguyen2023toward, le2024social} and Inverse Reinforcement Learning (IRL) \cite{park2020social, wang2021decision, phan2023driveirl} have gained traction for interactive navigation, capitalizing on the increasing availability of human-trajectory datasets. Despite their ability to mimic social norms, these data-driven approaches are frequently constrained by their training distributions; they often exhibit poor generalization or failure when encountering out-of-distribution (OOD) scenarios that deviate from the expert demonstrations.

\subsection{Quadrupedal Locomotion}
Traversing uneven terrain typically necessitates the generation of elevation maps \cite{agrawal2022vision, miki2022elevation, miki2022learning, kurniawan2025learning} to represent the topography beneath and surrounding the quadruped robot. However, maintaining high-frequency updates for these maps is often precluded by limited onboard computational resources. Moreover, map quality is frequently compromised by noise and motion blur resulting from oscillatory locomotion dynamics. To develop locomotion policies independent of explicit elevation maps, the teacher-student paradigm with a two-stage learning framework \cite{kim2024not, lee2020learning, lai2023sim, cheng2024quadruped, wu2025learn, mousa2025teacher} has emerged as a prominent solution. Furthermore, two-stage learning frameworks are employed to modulate foot placement atop foundational gaits, such as those governed by Central Pattern Generators (CPGs) \cite{seto2025learning}.

Rather than employing a two-stage training pipeline, single-stage asymmetric actor–critic (AC) techniques have become a cornerstone of sim-to-real RL. A notable example is Dream Walking for Quadrupedal robots (DreamWaQ) \cite{nahrendra2023dreamwaq}, where the critic is augmented with privileged information during training, while the actor operates exclusively on proprioceptive states to ensure feasible hardware deployment. Conversely, some approaches dispense with privileged information entirely by adopting symmetric AC learning, as seen in Self-learning Latent Representation (SLR) \cite{chen2024slr}. In this work, we extend the DreamWaQ architecture for our locomotion module. By leveraging asymmetric observations, we eliminate the need for explicit exteroceptive perception, thereby facilitating a direct and efficient sim-to-real transfer.

\begin{figure*}[!t]
	\centering
	\small
	\includegraphics[width=17.5cm]{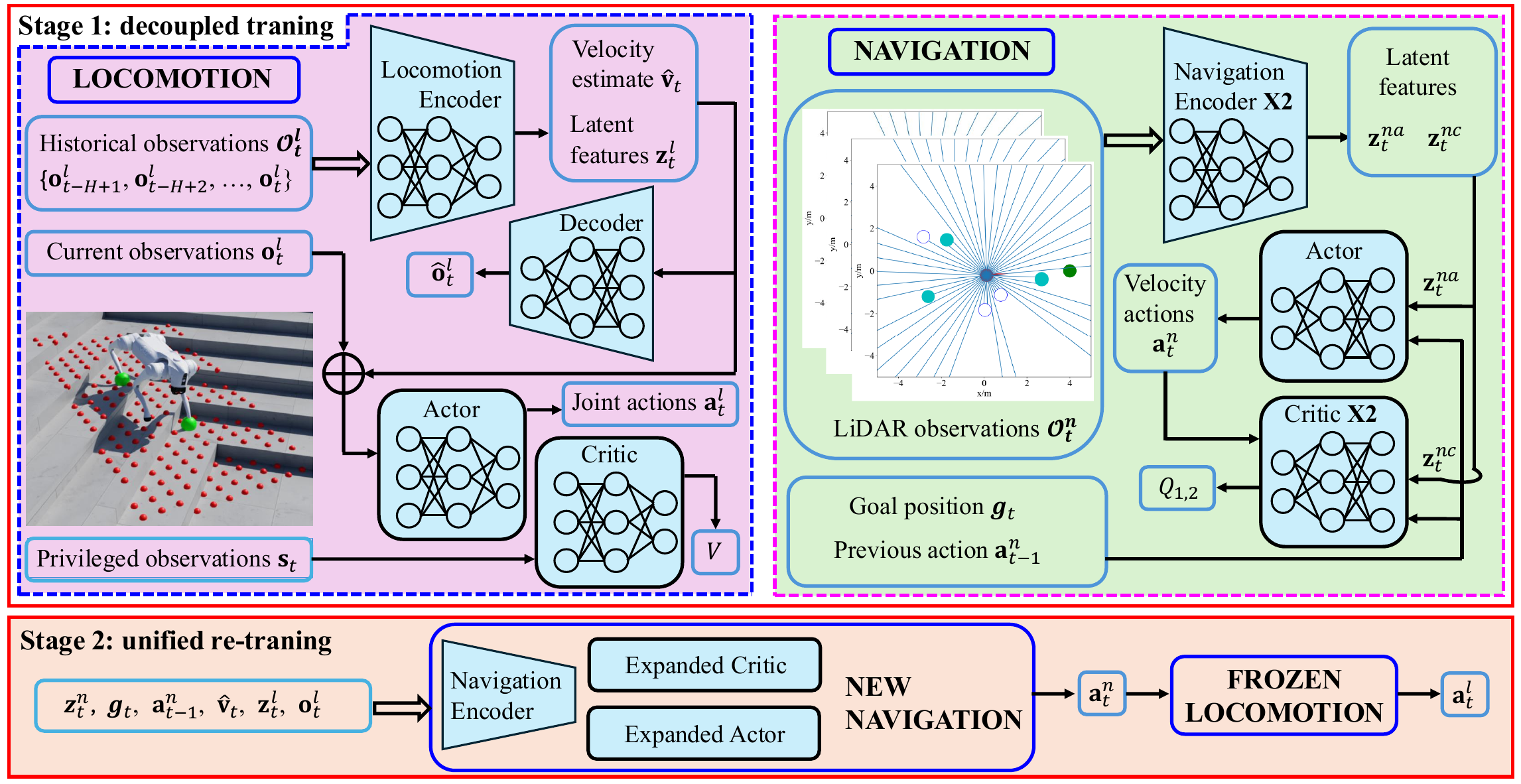}
	\caption{Overall framework with a two-stage training architecture. In the first stage, a quadrupedal locomotion policy and a navigation policy with interaction awareness are trained separately. In the second stage, these two policies are unified into a single quadrupedal navigation policy, which is further retrained in uneven and interactive environments.}
	\label{overallFramework}
\end{figure*}

\subsection{Quadrupedal Navigation}
Despite the maturity of wheeled mobile robot navigation, legged navigation particularly in dynamic settings, remains an open challenge due to the intricacies of high-dimensional robot dynamics and irregular terrain. Prevailing methodologies generally decouple legged navigation into a two-tier hierarchy \cite{xie2024real, mastalli2020motion, dudzik2020robust, hoeller2021learning, xiong2022model, liao2023walking, kim2023armp, feng2023gpf, seo2023learning, unlu2024control}: a high-level planner generates a feasible Center-of-Mass (CoM) trajectory, which a low-level whole-body controller (WBC) is then tasked to track. However, these hierarchical frameworks often struggle to maintain consistency between the planning and execution layers, as they fail to fully incorporate low-level locomotion constraints and the geometric nuances of uneven terrain when planning the CoM trajectory.

End-to-end quadrupedal navigation pipelines \cite{zhang2024resilient, imai2022vision, rudin2022advanced, bellegarda2024visual, zhang2024traversability} seek to unify locomotion, terrain adaptation, and collision avoidance with exteroceptive data and navigation-based objectives. Despite their conceptual elegance, these frameworks often encounter practical bottlenecks. Specifically, the update frequency of exteroceptive perception typically cannot match the high-rate requirements of locomotion control. Furthermore, high-frequency perceptual processing often introduces significant sensor noise and latency. In addition, a reliance on static, pre-built elevation maps restricts the robot’s adaptability to unexplored or dynamic environments.

To mitigate the dependence on high-frequency perception while accounting for locomotion constraints, frameworks such as Vision and Proprioception for Navigation (VP-Nav) \cite{fu2022coupling} first predict potential collisions and falls based on proprioceptive states, subsequently leveraging these predictions for low-frequency navigation tasks. Similarly, Visual Navigation and Locomotion (ViNL) \cite{kareer2023vinl} employs a teacher–student paradigm to distill exteroceptive data into a unified policy. Despite these advancements, current research in quadrupedal navigation is largely restricted to static environments, where latent environmental features are relatively stable and easy to extract. However, the most compelling applications for quadrupedal robots involve human-centric tasks, such as last-mile delivery and autonomous surveillance, which necessitate agile, reactive interactions with dynamic pedestrians across irregular terrains, including staircases, inclines, and unpaved surfaces.

\section{Framework and Problem Formulation} \label{overall_framework}

\subsection{Overall Framework}
The proposed framework, illustrated in Fig. \ref{overallFramework}, utilizes a two-stage training architecture comprising three DRL modules. During the first stage, two specialized policies are developed in parallel:

\textbf{Locomotion controller}: This controller is trained using high-fidelity and full-body quadruped models in IsaacSim \cite{mittal2023orbit} to ensure robust locomotion across uneven terrains.

\textbf{Navigation policy}: This policy is trained in interactive environments using a simplified CoM model governed by differential kinematics and associated kinematics constraints.

In the second stage, these policies are integrated into a unified architecture. This combined pipeline is then fine-tuned as the quadruped robot learns to navigate challenging terrains while simultaneously performing reactive collision avoidance with dynamic pedestrians and static obstacles.

\subsection{Problem Formulation}
We formulate quadrupedal locomotion, navigation with interaction awareness, and integrated quadrupedal navigation in uneven, interactive environments as Partially Observable Markov Decision Processes (POMDPs). A POMDP is defined by the tuple $\mathcal{P} = \langle S, A, T, R, \Omega, O, \gamma \rangle$, where $S$ denotes the state space, $A$ the action space, $T(s'|s \in S, a \in A)$ the probability of transitioning to state $s'$ from state $s$ after taking action $a$, $R(s,a)$ the reward for taking action $a$ in state $s$, $\Omega$ the observation space, $O(o|s', a)$ the probability of receiving observation $o$ after taking action $a$ and arriving in state $s'$, and $\gamma$ the reward discount factor.

To solve these POMDPs without explicit knowledge of the underlying transition and observation models $(T, O)$, we employ model-free RL to iteratively optimize the policy through environmental interaction. We define the interaction between the agent and the environment as follows:
$$
s_t \in S \xrightarrow{a_t \in A} s_{t+1} \in S, r_t \in R.
$$
The agent aims to learn a policy $\pi(a_t|s_t)$ that maximizes the expected cumulative reward:
\begin{equation}
    J(\pi)=\mathbb{E}_{\pi}\left[\sum_{t=0}^{\infty} \gamma ^t r_t \right].
    \label{problem_formulation}
\end{equation}
We employ a suite of specialized DRL algorithms to optimize the policies across our multi-stage framework, encompassing quadrupedal locomotion, navigation with interaction awareness, and their subsequent integration.

To maintain consistency across our multi-task framework, we adopt a standardized notation. Since the three DRL pipelines share a common POMDP formulation, we utilize superscripts $\{l, n, u\}$ to distinguish between locomotion, navigation with interaction awareness, and unified navigation in uneven and interactive environments, respectively. Furthermore, within the Actor–Critic (AC) framework, we employ superscripts $\{a, c\}$ to differentiate between the actor and critic state spaces. Under this convention, for example, $s_t^{u,c}$ represents the state of the critic in the unified navigation module at time $t$.

\section{Quadrupedal Locomotion} \label{quadruped_locomotion}
Our locomotion controller, TRAIN-Loco, builds upon the DreamWaQ \cite{nahrendra2023dreamwaq} architecture. While maintaining the core DRL workflow, we significantly enhance the system by redesigning the encoder network and refining the reward structures to improve stability and gait efficiency. 

\subsection{Observations and States}
Precise knowledge of terrain geometry and full-body contact dynamics is essential for stable quadrupedal locomotion on irregular surfaces. However, real-world exteroceptive sensing is inherently noisy and requires substantial computational overhead to maintain high-frequency, real-time elevation mapping. Furthermore, a strict reliance on pre-built terrain maps compromises the robot's adaptability to unforeseen or dynamic environments. To circumvent these limitations, we adopt an asymmetric actor–critic architecture \cite{pinto2018asymmetric}. This framework allows the critic to leverage privileged terrain and contact data during training, while the actor learns to infer these implicit environmental features solely from proprioceptive history, ensuring robust deployment on hardware without the need for explicit mapping.

During the training phase, the critic network has access to the full simulation state, including both proprioceptive observations $\mathbf{o}_t^l$ and privileged exteroceptive information. Since the critic is discarded prior to real-world deployment, this asymmetric information flow allows it to provide more accurate value estimations during policy optimization. The proprioceptive observation vector, $\mathbf{o}_t^l \in \mathbb{R}^{45}$, provides the robot with its current kinematic and dynamic status and is defined as: 
\begin{equation}
    \mathbf{o}_t^l = \left[\boldsymbol{\omega}_t, \mathbf{g}_t, \mathbf{c}_t, \boldsymbol{\theta}_t, \boldsymbol{\dot{\theta}}_t, \mathbf{a}_{t-1}^l\right],
    \label{observation}
\end{equation}
where $\boldsymbol{\omega}_t \in \mathbb{R}^3$ represents the base angular velocity, $\mathbf{g}_t \in \mathbb{R}^3$ is the gravity vector projected into the robot's base frame, $\mathbf{c}_t \in \mathbb{R}^3$ denotes the target velocity command provided by the high-level planner, $\boldsymbol{\theta}_t \in \mathbb{R}^{12}$ and $\boldsymbol{\dot{\theta}}_t \in \mathbb{R}^{12}$ respectively correspond to the joint positions and velocities, and $\mathbf{a}_{t-1}^l \in \mathbb{R}^{12}$ are the angular offsets relative to the default standing configuration from the previous time step, included to promote temporal smoothness in control.

\begin{figure}[!t]
	\centering
	\small
	\includegraphics[width=8.5cm]{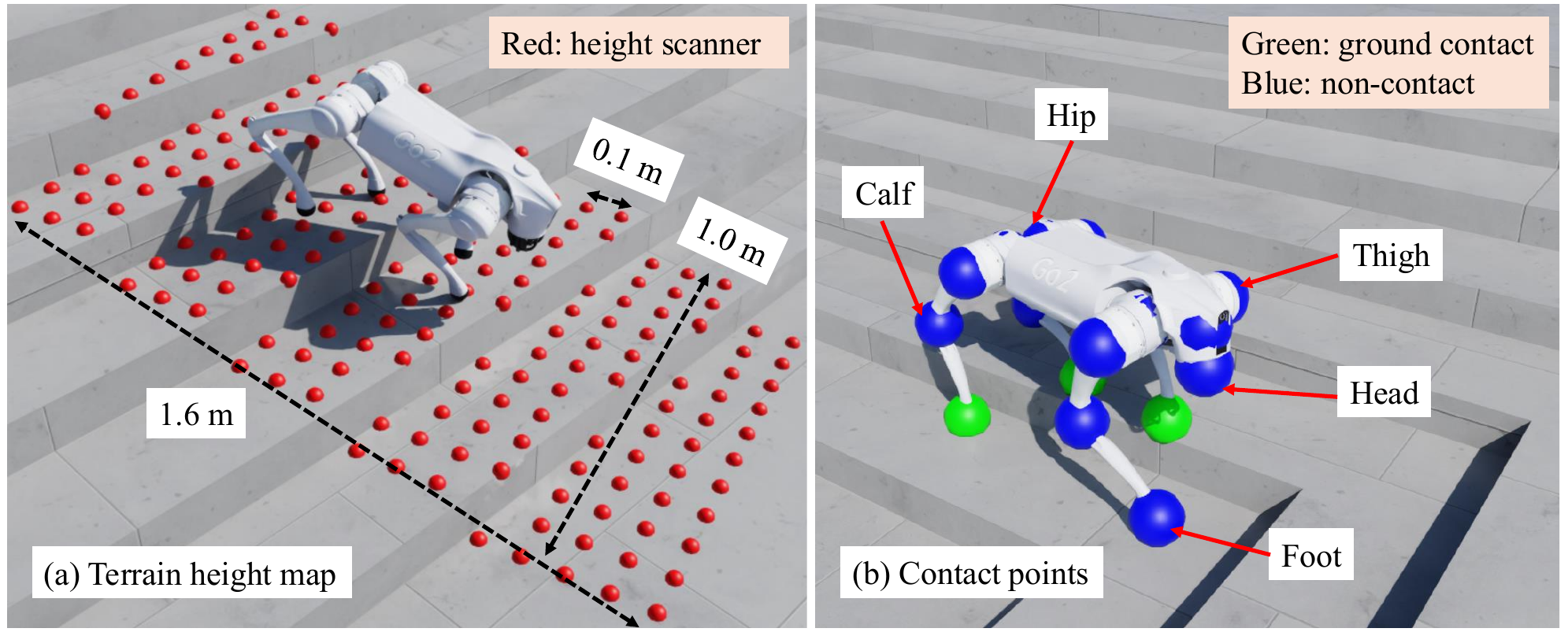}
	\caption{Terrain height map and contact points visualized in IsaacSim.}
	\label{terrainContact}
\end{figure}

The exteroceptive component of the privileged state incorporates a local terrain height map, $\boldsymbol{\mathcal{H}}_t \in \mathbb{R}^{187}$, and full-body contact forces, $\boldsymbol{\mathcal{C}}_t \in \mathbb{R}^{51}$. As illustrated in Fig. \ref{terrainContact}(a), the height map is sampled over a $1.6 \times 1.0$ m grid centered on the robot with a resolution of $0.1$ m. To capture comprehensive interaction dynamics, we monitor 17 distinct body contact points, as depicted in Fig. \ref{terrainContact}(b). These points include the four feet, calves, thighs, and hips, alongside the head. For each contact point, the critic receives the three-dimensional force components ($x, y, z$) in the robot frame.

To account for the inherent noise and estimation drift associated with the body linear velocity $\mathbf{v}_t \in \mathbb{R}^3$ on physical hardware, we exclude it from the actor's observation vector $\mathbf{o}_t^l$. Instead, we treat $\mathbf{v}_t$ as a privileged component available only during training. This ensures the locomotion policy remains robust to velocity estimation errors while allowing the critic to utilize the ground-truth velocity for accurate value estimation. In summary, the comprehensive critic state is defined as:
\begin{equation}
    \mathbf{s}_t^{l,c} = \left[\mathbf{o}_t^l, ~\boldsymbol{\mathcal{H}}_t, ~\boldsymbol{\mathcal{C}}_t, ~\mathbf{v}_t\right] \in \mathbb{R}^{286}.
    \label{critic_states}
\end{equation}

To ensure reliable and efficient real-world deployment, the actor network, which constitutes the final onboard locomotion policy, excludes the privileged variables $\boldsymbol{\mathcal{H}}_t$, $\boldsymbol{\mathcal{C}}_t$, and $\mathbf{v}_t$. Instead, as depicted in Fig. \ref{overallFramework}, the actor relies on a temporal encoder to extract a latent representation $\mathbf{z}^l_t \in \mathbb{R}^{16}$ that implicitly encodes terrain geometry and contact dynamics. Simultaneously, the encoder provides an estimate of the body linear velocity, $\hat{\mathbf{v}}_t \in \mathbb{R}^3$. These features are derived from a historical observation buffer, $\boldsymbol{\mathcal{O}}_t^l$, defined as:
\begin{equation}
    \boldsymbol{\mathcal{O}}_t^l = \left[\mathbf{o}_{t-H+1}^l, \mathbf{o}_{t-H+2}^l, \dots, \mathbf{o}_t^l\right] \in \mathbb{R}^{10 \times 45},
    \label{history_observation_locomotion}
\end{equation}
where the history length is set to $H = 10$. Consequently, the actor state, $\mathbf{s}_t^{l,a}$, is formulated as:
\begin{equation}
    \mathbf{s}_t^{l,a} = \left[\mathbf{o}_t^l, ~\mathbf{z}^l_t, ~\hat{\mathbf{v}}_t\right] \in \mathbb{R}^{64}.
    \label{actor_states}
\end{equation}
By utilizing $\mathbf{s}_t^{l,a}$, the actor can effectively navigate complex environments by inferring critical physical properties from its own motion history, bypassing the need for high-frequency exteroceptive mapping.

\subsection{Locomotion Actions}
The locomotion actions are defined as $\mathbf{a}_t^l \in \mathbb{R}^{12}$, representing the angular offsets relative to the default standing configuration, $\boldsymbol{\theta}_{\mathrm{stand}}$. To ensure mechanical safety and maintain gait stability, the absolute magnitude of each action component is capped at $0.25~\text{rad}$. The target joint positions, $\boldsymbol{\theta}_{\mathrm{des}}$, are calculated as:
\begin{equation}
    \boldsymbol{\theta}_{\mathrm{des}} = \boldsymbol{\theta}_{\mathrm{stand}} + \mathbf{a}_t^l.
    \label{locomotion_desired_joint}
\end{equation}

To track these setpoints, a low-level Proportional-Derivative (PD) controller computes the required joint torques $\boldsymbol{\tau}_t$:
\begin{equation}
    \boldsymbol{\tau}_t = \mathbf{K}_p (\boldsymbol{\theta}_{\mathrm{des}} - \boldsymbol{\theta}_t) - \mathbf{K}_d \boldsymbol{\dot{\theta}}_t,
    \label{jointPDControl}
\end{equation}
where $\mathbf{K}_p$ and $\mathbf{K}_d$ denote the diagonal matrices of proportional and derivative gains, respectively, and $\boldsymbol{\dot{\theta}}_t$ represents the current joint velocities. This hierarchical control structure enables the RL policy to operate at a lower frequency while the PD controller ensures high-frequency tracking and joint impedance.

\subsection{Networks}

As illustrated in Fig. \ref{overallFramework}, the TRAIN-Loco architecture is composed of three primary modules: the actor network, the critic network, and a temporal auto-encoder. The specifications for these networks are detailed in APPENDIX \ref{locomotionNetworkDetail}.

\textbf{Actor and critic networks}: Both utilize Multi-Layer Perceptrons (MLPs) with an identical hidden layer configuration to maintain representational parity. Their architectures diverge only at the output stage to accommodate their respective tasks. The actor network, or policy $\pi_{\phi^l}(\mathbf{a}_t^l \mid \mathbf{s}_t^{l,a})$, infers the joint action $\mathbf{a}_t^l$, while the critic network $V_{\psi^l}(\mathbf{s}_t^{l,c})$ estimates the state-value function for policy evaluation.

\textbf{Temporal auto-encoder}: This module serves as the core for latent feature extraction. The encoder employs a hybrid structure of Convolutional Neural Networks (CNNs) and MLPs to process the historical observation buffer $\boldsymbol{\mathcal{O}}_t^l$, while the decoder consists of a two-layer MLP designed to reconstruct the proprioceptive state $\mathbf{\hat{o}}_t^l$. We employ a variational temporal encoder $f_{\varphi^l}(\boldsymbol{\mathcal{O}}_t^l)$ to extract the estimated body linear velocity $\hat{\mathbf{v}}_t^l$ alongside latent terrain and contact features $\mathbf{z}_t^l$. Subsequently, a decoder network $f_{\rho}(\mathbf{z}_t^l, \hat{\mathbf{v}}_t^l)$ reconstructs the current observation as $\hat{\mathbf{o}}_t^l$. The network parameters are updated jointly via a reconstruction loss between $\mathbf{\hat{o}}_t^l$ and the ground-truth observation $\mathbf{o}_t^l$. Furthermore, an estimation loss is formulated using $\hat{\mathbf{v}}_t^l$ and the privileged linear velocity data provided by the simulator.

\textbf{Activation and layers}: All hidden layers across the three networks employ the Exponential Linear Unit (ELU) activation function to ensure smooth gradient flow. In contrast, the output layers remain linear to allow for unconstrained regression of actions, values, and reconstructed observations.

The network parameters $\{\phi^l, \psi^l, \varphi^l, \rho\}$ are optimized via a multi-objective loss function. This objective integrates the Proximal Policy Optimization (PPO) loss \cite{schulman2017proximal} for locomotion control, a mean square error (MSE) term for velocity estimation, and a Variational Autoencoder (VAE) loss \cite{kingma2013auto} to facilitate representation learning.

\subsection{Rewards}

\renewcommand\arraystretch{1.3}
\begin{table}[!t]
	\small
	\centering
	\caption{Reward function for quadrupedal locomotion}
	\begin{tabular}{c@{\hspace{1pt}} c@{\hspace{1pt}} c} 
		\hline
		Reward                & Equation ($r_i$)  & Weight ($w_i$)   \\
		\hline 
		  Linear $\mathrm{vel}$ track     & $\exp \{-4\left(\mathbf{v}_{xy}^{\mathrm{cmd}} - \mathbf{v}_{xy}\right)^2\}$   & $1.5$     \\
            Angular $\mathrm{vel}$ track    & $\exp \{-4\left(w_{z}^{\mathrm{cmd}} - w_{z}\right)^2\}$ & $0.75$    \\
            Linear $\mathrm{vel}$ ($z$)     & $v_z^2$                                                                        & $-2.0$     \\
            Angular $\mathrm{vel}$ ($xy$)   & $\boldsymbol{w}_{xy}^2$                                                        & $-0.05$    \\
            Torque                & $ \boldsymbol{{\tau}} ^2$                                                   & $-2 \times 10^{-4}$    \\
            Angular acceleration      & $\boldsymbol{\ddot{{\theta}}} ^2$ & $-2.5 \times 10^{-7}$ \\
            Action rate           & $\left(\mathbf{a}_t^l - \mathbf{a}_{t-1}^l \right) ^2$& $-0.01$ \\
            Feet air time         & $ \left(\mathbf{t}_a^{\mathrm{last}} - 0.5\right) \cdot \mathbf{t}_c^{\mathrm{check}} \cdot ||\mathbf{v}_{xy}^\mathrm{cmd}||_2$       & $0.01$ \\
            \hline
	\end{tabular}
	\label{tableRewardLocomotion}
\end{table}

The reward structure for TRAIN-Loco builds upon the DreamWaQ framework \cite{nahrendra2023dreamwaq}, with refinements specifically engineered to improve learning efficiency and cross-platform generalizability. As summarized in TABLE \ref{tableRewardLocomotion}, the total reward consists of task rewards, which penalize deviations from the commanded velocity, and stability rewards, which promote natural gait patterns and structural safety. 

In our formulation, the operator $\exp(\cdot)$ denotes the exponential function, while superscripts $(\cdot)^{\text{cmd}}$ distinguish target values from measured states. All kinematic variables are defined in the robot's local body frame, where the $x$-axis and $z$-axis point forward and upward, respectively. Specifically:

\begin{itemize}
    \item $\mathbf{v}_{xy}$ and $\boldsymbol{w}_{xy}$ represent the linear and angular velocities in the horizontal plane. The commands for $\mathbf{v}_{xy}$ are randomly set from -1.0 to 1.0 m/s during training.

    \item $v_z$ and $w_{z}$ denote the vertical linear velocity and yaw rate. The command for $w_{z}$ ranges from -1.0 to 1.0 rad/s.

    \item $\boldsymbol{\tau}$ and $\boldsymbol{\ddot{\theta}}$ correspond to the applied joint torques and joint accelerations, respectively.
\end{itemize}

To regulate gait timing and contact dynamics, we utilize $\mathbf{t}_a^{\text{last}}$ to denote the flight duration prior to the most recent contact event. Additionally, the boolean operator $\mathbf{t}_c^{\text{check}}$ determines whether a foot has established ground contact within the preceding control interval $\Delta t^l = 0.02$ s ($50$ Hz).

The total reward is computed as:
\begin{equation}
    r_t^l\left(\mathbf{s}_t^{l,c},\mathbf{a}_t^l\right)=\sum{r_i \cdot w_i \cdot \Delta t^l},
    \label{rewardLocomotion}
\end{equation}
where $i$ is the index of each reward shown in TABLE \ref{tableRewardLocomotion}.

\subsection{Curriculum Learning and Domain Randomization}
The training environment encompasses a diverse range of challenging topographies, including flat ground, inclines, stairs, rough terrain, and stepped grids, as illustrated in Fig. \ref{introductionFigure}. The geometric complexity is parameterized as follows:
\begin{itemize}
    \item Slopes: Angles ranging from $0^{\circ}$ to $23^{\circ}$.

    \item Stairs: Step heights varying within $[0.05, 0.12]$ m.

    \item Rough surfaces: Stochastic height noise ranging from $0.01$ to $0.06$ m.

    \item Stepped grids: Discrete height variations between $0.025$ and $0.1$ m.
\end{itemize}

To ensure stable policy convergence across these diverse conditions, we employ curriculum learning to adaptively modulate terrain difficulty. The curriculum operates on a performance-based heuristic: the terrain level is incremented upon successful and stable traversal of the current difficulty, while a failure to maintain stability triggers a reduction in difficulty. This progressive approach allows the agent to build foundational gait patterns before tackling the most extreme environmental constraints.

\renewcommand\arraystretch{1.2}
\begin{table}[!t]
	\small
	\centering
	\caption{Locomotion Domain Randomization}
	\begin{tabular}{c c c} 
		\hline
		Type  & Domain                      & Noise         \\
		\hline 
        \multirow{4}{*}{System dynamics}    & Friction coefficient     & [0.65, 1.25]      \\
		                                      & Body mass                & [-1.0, 3.0] kg     \\
                                            & CoM $xy$                 & [-0.05, 0.05] m     \\
                                            & CoM $z$                  & [-0.01, 0.01] m      \\
        \hline
        \multirow{6}{*}{Observations} & Body linear $\mathrm{vel}$ $\mathbf{v}_t$  & [-0.1, 0.1] m/s          \\
                                      & Body angular $\mathrm{vel}$ $\boldsymbol{\omega}_t$ & [-0.2, 0.2] rad/s        \\
                                      & Projected gravity $\mathbf{g}_t$          & [-0.1, 0.1] m              \\
                                      & Joint position $\boldsymbol{\theta}_t$              & [-0.01, 0.01] rad        \\
                                      & Joint velocity $\boldsymbol{\dot{\theta}}_t$            & [-1.5, 1.5] rad/s        \\
                                      & Height map $\boldsymbol{\mathcal{H}}_t$                 & [-0.1, 0.1] m            \\
        \hline
        Reset                         & Initial joint offset             & [-0.1, 0.1] rad \\
		\hline
	\end{tabular}
	\label{locomotionDomainRandomization}
\end{table}

Despite the fidelity of modern simulators, significant discrepancies between simulated environments and the physical world, termed the sim-to-real gap, can compromise the deployment of locomotion policies on real hardware. To bridge this gap, we employ Domain Randomization (DR), a technique that exposes the policy to a wide distribution of physical parameters during training.

Our DR implementation, detailed in TABLE \ref{locomotionDomainRandomization}, is modeled after established benchmarks in robust locomotion \cite{miki2022learning, nahrendra2023dreamwaq, chen2024slr}. We introduce stochasticity by sampling from uniform distributions and applying these perturbations to:

\begin{itemize}
    \item System dynamics: Randomizing default robot parameters such as body mass, center of mass (CoM) positions, and ground friction coefficients.

    \item Observations: Injecting noise into proprioceptive and terrain height measurements.

    \item Reset: Randomizing initial joint angles to start from diverse initial robot states.
\end{itemize}

\begin{algorithm}
\caption{TRAIN-Loco}
\label{trans_loco}
\begin{algorithmic}[1]
\STATE \textit{Initialize}: $\mathrm{itr} \leftarrow 1$, randomize masses and CoMs for $N_r$ quadruped robots in IsaacSim
\STATE \textit{Reset}: Robot pose in the world frame, terrain level, PPO episode buffer $\mathcal{D}$, $\mathrm{reset}$ $\leftarrow$ \textit{False}, $t \leftarrow 0$
\STATE \textit{Observe}: $\mathbf{o}_t^l$, $\mathbf{s}_t^{l,c}$, $\boldsymbol{\mathcal{O}}_t^l$
\WHILE{$\mathrm{itr} \leq \mathcal{I}$}
    \STATE Compute $\hat{\mathbf{v}}_t$ and $\mathbf{z}_t^l$ with $f_{\varphi^l}(\boldsymbol{\mathcal{O}}_t^l)$
    \STATE Construct $\mathbf{s}_{t}^{l,a}$ with $\mathbf{o}_t^l$, $\hat{\mathbf{v}}_t$ and $\mathbf{z}_t^l$
    \STATE Sample an action $\mathbf{a}_{t}^l$ with $\pi_{\phi^l}(\mathbf{s}_t^{l,a})$
    \STATE Execute one step with the action $\mathbf{a}_{t}^l$ in IsaacSim
    \STATE $t \leftarrow t+1$
    \STATE \textit{Observe}
    \STATE Calculate $r_t^l$ using Eq. (\ref{rewardLocomotion})
    \STATE Estimate the value function $V_{\psi ^l}(\mathbf{s}_t^{l,c})$
    \IF{robot fall \OR episode timeout}
        \STATE $\mathrm{reset}$ $\leftarrow$ \textit{True}
    \ENDIF
    \STATE Fill $\mathcal{D}$ with $r_t^l$, $\mathbf{a}_{t}^l$, $\mathbf{o}_t^l$, $\mathbf{s}_t^{l,c}$, $\boldsymbol{\mathcal{O}}_t^l$, $V_{\psi ^l}(\mathbf{s}_t^{l,c})$ and $\mathrm{reset}$
    \IF{$\mathrm{reset}$}
        \STATE Sample batches from $\mathcal{D}$
        \STATE Update network parameters $\phi ^l$, $\psi ^l$, $\varphi ^l$, and $\rho$ via PPO \cite{schulman2017proximal} and VAE \cite{kingma2013auto}
        \STATE Update terrain level via curriculum learning
        \STATE \textit{Reset}
        \STATE \textit{Observe}
        \STATE $\mathrm{itr}$ $\leftarrow$ $\mathrm{itr} + 1$
    \ENDIF
\ENDWHILE
\end{algorithmic}
\end{algorithm}

In summary, the locomotion framework, TRAIN-Loco, is illustrated in \textbf{Algorithm} \ref{trans_loco}.

\section{Navigation with Interaction Awareness} \label{social_navigation}
In parallel with TRAIN-Loco, we propose a straightforward framework for navigation with interaction awareness, termed TRAIN-Nav. This module directly maps transformed LiDAR scans to robot actions while adhering to the velocity constraints of differential-drive kinematics.

\subsection{Navigation Actions and Kinematics}
\textbf{Actions}. To simplify the navigation task and improve stability, we model the quadruped robot as a circular ego-agent with radius $r^{\mathrm{agent}}$. The agent's movement is governed by differential-drive kinematics, with linear velocity $v^n \in [0, v_{\max}]$ and angular velocity $w^n \in [-w_{\max}, w_{\max}]$. Notably, we exclude lateral ($y$-direction) velocity commands; our empirical observations indicated that lateral movement frequently caused leg self-collisions, particularly when traversing challenging terrains. Accordingly, the navigation action space is defined as:
\begin{equation}
    \mathbf{a}_t^n = [v_t^n, w_t^n]^T.
    \label{actionNavigation}
\end{equation}
We set $v_{\max}=0.5$ m/s and $w_{\max}=1.0$ rad/s.

\textbf{Differential kinematics}. Given the action $\mathbf{a}_t^n$ and the current ego-agent state $\mathbf{x}_t = [x_t, y_t, \theta_t^n]^T$, where $(x_t, y_t)$ and $\theta_t^n$ represent the position and orientation in the 2D global frame, the state is updated according to the differential-drive kinematic model:
\begin{equation}
        \mathbf{x}_{t+1}=\mathbf{x}_t + \mathbf{B} \mathbf{a}_t^n \Delta t^n,
    \label{differentialKinematics}
\end{equation}
where $\Delta t^n = 0.2$ s is the motion planning period for navigation, and $\mathbf{B}$ denotes the control matrix:
\[
\mathbf{B} =
\begin{bmatrix}
\cos{\theta _t ^n} & 0 \\
\sin{\theta _t ^n} & 0 \\
0 & 1
\end{bmatrix}.
\]

\textit{Remark 1: We employ differential-drive kinematics for navigation in the initial training stage rather than modeling full-body quadrupedal dynamics. This abstraction offers high computational efficiency and provides a ``warm start'' for the subsequent stage, significantly accelerating overall learning. Additionally, this kinematic simplification facilitates direct comparisons between our learning-based policies and a variety of traditional model-based motion planners.}

\subsection{Observations and States}

\begin{figure}[!t]
	\centering
	\small
	\includegraphics[width=8.5cm]{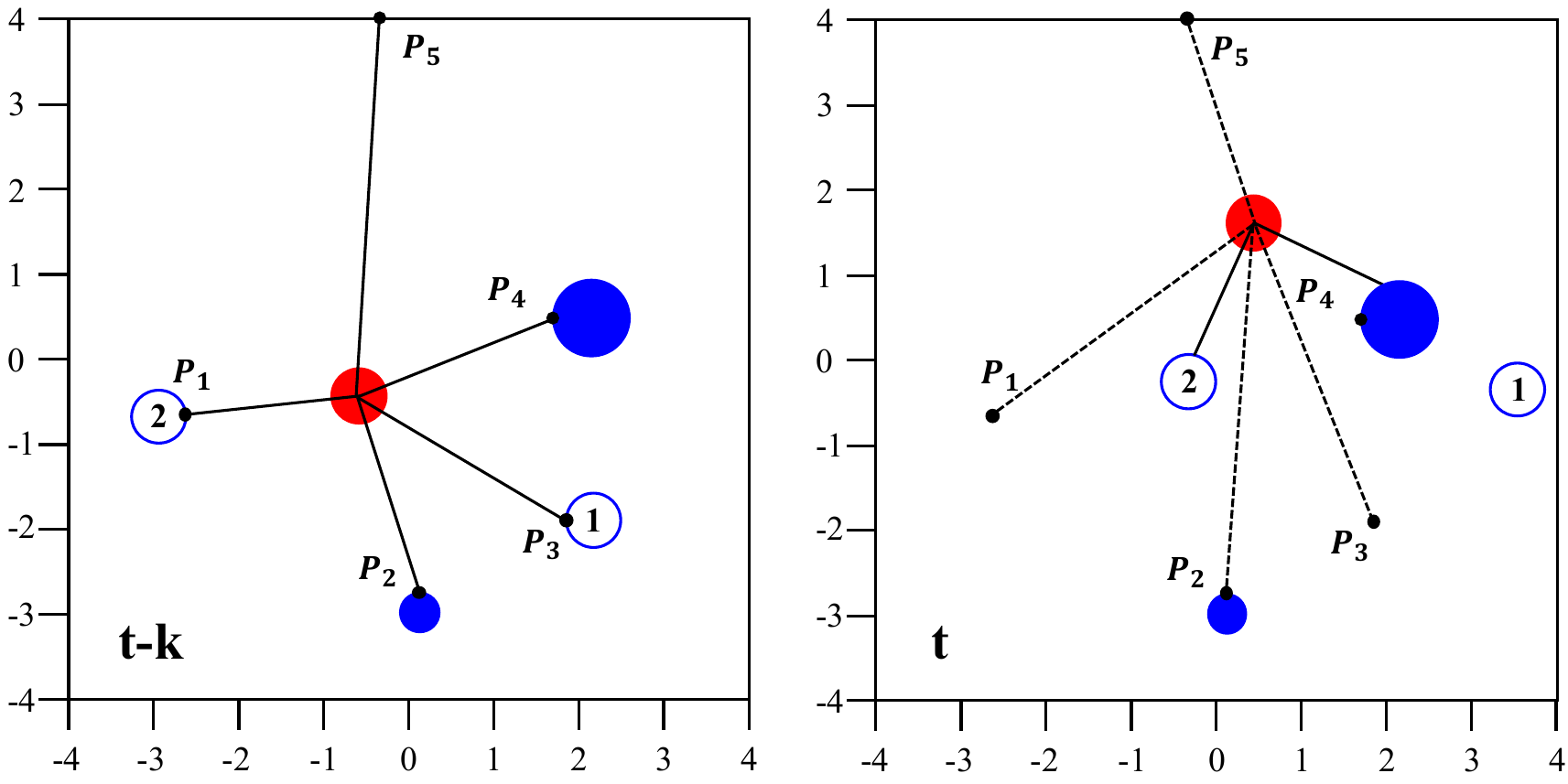}
	\caption{LiDAR scan transformation. The red filled circle represents the ego-agent, blue filled circles denote static obstacles of varying sizes, and blue hollow circles indicate pedestrians with a fixed radius. The left panel shows the LiDAR scan (solid beams) at time $t-k$, while the right panel illustrates the scan (solid beams) at the current time $t$. Dashed lines in the right panel correspond to the transformed LiDAR beams from $t-k$ to $t$. Each point $P_i$ retains the same global position in both panels.}
	\label{LidarTransform}
\end{figure}

To capture the complex dynamics of pedestrian and obstacle interactions, we introduce a novel representation termed \textit{transformed LiDAR scans}. As shown in Fig. \ref{LidarTransform}, the raw LiDAR observation $\mathbf{o}_t^n \in \mathbb{R}^{1800}$ provides a $360^{\circ}$ field of view. To incorporate temporal context, we maintain a buffer of historical scans. For a scan captured at time $t-k$, the local position of each interaction point $P_i$ in the robot’s coordinate frame is denoted as $\mathbf{p}_{t-k}^{i\text{-}\mathrm{loc}} = [x_{t-k}^{i\text{-}\mathrm{loc}},\, y_{t-k}^{i\text{-}\mathrm{loc}}]^T$:
\begin{equation}
    \begin{split}
        \vartheta _{t-k} ^i &= 2\pi \cdot \frac{i}{N}, i \in \{1, 2, ..., n\},\\
        x_{t-k}^{i \text{-} \mathrm{loc}} &= r_{t-k}^i  \cos \vartheta _{t-k} ^i, ~y_{t-k}^{i \text{-} \mathrm{loc}} = r_{t-k}^i  \sin \vartheta _{t-k} ^i,\\
    \end{split}
    \label{pointToLocal}
\end{equation}
where $N=1800$ denotes the number of LiDAR beams, $\vartheta_{t-k}^i$ is the beam angle, and $r_{t-k}^i$ is the corresponding range measurement. At time $t-k$, given the global position and orientation of the ego-agent $\mathbf{p}_{t-k}^{\mathrm{ego}} = [x_{t-k},\, y_{t-k}]^T$ and $\theta_{t-k}^n$, we compute the global position of each interaction point $P_i$, denoted as $\mathbf{p}_{t-k}^{i\text{-}\mathrm{glob}} = [x_{t-k}^{i\text{-}\mathrm{glob}},\, y_{t-k}^{i\text{-}\mathrm{glob}}]^T$:
\begin{equation}
    \mathbf{p}_{t-k}^{i \text{-} \mathrm{glob}} = \mathbf{p}_{t-k}^{\mathrm{ego}} + \mathbf{R}(\theta _{t-k}^n)\mathbf{p}_{t-k}^{i \text{-} \mathrm{loc}},
    \label{pointToGlobal}
\end{equation}
where $\mathbf{R}(\theta _{t-k}^n)$ is the rotation matrix:
\[
\mathbf{R}(\theta _{t-k}^n) =
\begin{bmatrix}
\cos{\theta _{t-k}^n} & -\sin{\theta _{t-k}^n} \\
\sin{\theta _{t-k}^n} & \cos{\theta _{t-k}^n} \\
\end{bmatrix}.
\]

Given the current ego-agent position and orientation $\mathbf{p}_{t}^{\mathrm{ego}} = [x_t,\, y_t]^T$ and $\theta_t^n$ on the global frame, the position of $P_i$ on the current robot frame can be obtained by inverting Eq.~(\ref{pointToGlobal}):
\begin{equation}
    \tilde{\mathbf{p}}_{t-k}^{i \text{-} \mathrm{loc}} = \mathbf{R}^{-1}(\theta _{t}^n)(\mathbf{p}_{t-k}^{i \text{-} \mathrm{glob}} - \mathbf{p}_{t}^{\mathrm{ego}}),
    \label{pointToCurrentRobot}
\end{equation}
where $\tilde{\mathbf{p}}_{t-k}^{i \text{-} \mathrm{loc}} = [\tilde{x}_{t-k}^{i \text{-} \mathrm{loc}}, \tilde{y}_{t-k}^{i \text{-} \mathrm{loc}}]^T$ denotes the position of $P_i$ on the current ego-agent frame. Finally, we reconstruct the LiDAR scan as:
\begin{equation}
        \tilde{r}_{t-k}^i = ||\tilde{\mathbf{p}}_{t-k}^{i \text{-} \mathrm{loc}}||_2,
    \label{transformedRange}
\end{equation}
where $\tilde{r}_{t-k}^i$ denotes the range from the ego-agent to $P_i$ at the current time step. Therefore, the transformed LiDAR scan is represented as: 
\begin{equation}
    \tilde{\mathbf{o}}_{t-k}^n = \{\tilde{r}_{t-k}^i|i=1, 2, ..., N\}^T.
    \label{reconstructedLidar}
\end{equation}

Together with the current LiDAR scan $\mathbf{o}_{t}^n$, we represent the final observation as:
\begin{equation}
    \boldsymbol{\mathcal{O}} _t ^n = \left[\tilde{\mathbf{o}}_{t-K+1}^n, ..., \tilde{\mathbf{o}}_{t-1}^n,  \mathbf{o}_{t}^n\right]\in \mathbb{R}^{6 \times1800},
    \label{observationNavigation}
\end{equation}
where $K=6$ is the total number of LiDAR scans, with the most recent scan kept in its original form and the previous $K-1$ scans transformed into the current ego-agent frame.  

By transforming historical LiDAR scans into the current robot frame, we filter out the ego-agent's motion, allowing the network to isolate and represent the independent movements of pedestrians. Furthermore, this transformation aids in identifying static obstacles, as their re-projected points remain spatially consistent across time steps. Our experiments, detailed in Section \ref{simulation_evaluations_navigation}-B, demonstrate that this representation significantly enhances navigation performance compared to the use of raw LiDAR scans.

Given the observations $\boldsymbol{\mathcal{O}}_t^n$, we extract latent states $\mathbf{z}_t^{n,a} \in \mathbb{R}^{50}$ and $\mathbf{z}_t^{n,c} \in \mathbb{R}^{50}$ for the actor and critic using two encoders with identical architectures. The navigation task is guided by the goal position in the robot frame, $\boldsymbol{g}_t = [r_t^{\mathrm{goal}}, \theta_t^{\mathrm{goal}}]$, where $r_t^{\mathrm{goal}}$ and $\theta_t^{\mathrm{goal}}$ represent the goal distance and relative orientation in the ego-agent frame, respectively. To account for reciprocal interactions, where the ego-agent’s movements influence pedestrian behavior, the previous action $\mathbf{a}_{t-1}^n$ is also included in the state representation. Therefore, the combined state is represented as:
\begin{equation}
    \mathbf{s}_t^{n,*} = \left[\mathbf{z}_t^{n,*}, \boldsymbol{g}_t,\mathbf{a}_{t-1}^n\right] \in \mathbb{R}^{54},~* \in \{a,c\}.
    \label{stateNavigation}
\end{equation}

In contrast to the locomotion module's use of asymmetric states, the state $\mathbf{s}_t^n$ for TRAIN-Nav is symmetric for both the actor and critic networks. This design choice stems from the fact that stable, low-frequency LiDAR scans are more readily accessible in real-world scenarios than precise terrain height maps or full-body contact data. Consequently, there is no requirement to distill privileged sensing data into the actor network for the navigation task.

\subsection{Networks}

The neural network architecture for TRAIN-Nav comprises three primary components: the actor and critic networks, implemented as MLPs, and the encoder network, which utilizes a combination of CNNs and MLPs. Comprehensive structural details are summarized in APPENDIX \ref{navigationNetworkDetail}. All hidden layers employ the Rectified Linear Unit (ReLU) activation function, while the output layers remain linear.

The actor network $\pi_{\phi^n}(\mathbf{a}_t^n|\mathbf{s}_t^{n})$ parameterizes a normal distribution by outputting the mean and log-variance of the action $\mathbf{a}_t^n$. Since we employ the Soft Actor-Critic (SAC) algorithm \cite{haarnoja2018soft}, the critic architecture differs from the locomotion module. Specifically, the critic estimates the action-value function $Q(\mathbf{s}_t^n, \mathbf{a}_t^n)$, taking both the state and action as inputs. To mitigate value overestimation, we utilize twin critics ($Q_{\psi_1^n}$ and $Q_{\psi_2^n}$) alongside two corresponding target critics ($Q_{\bar{\psi}_1^n}$ and $Q_{\bar{\psi}_2^n}$), which stabilize training by smoothing the bootstrapped targets. Latent features for the actor and critic, denoted as $\mathbf{z}_t^{n,a}$ and $\mathbf{z}_t^{n,c}$, are extracted from two encoder networks $f_{\varphi^{n,a}}(\boldsymbol{\mathcal{O}}_t^n)$ and $f_{\varphi^{n,c}}(\boldsymbol{\mathcal{O}}_t^n)$, respectively:
\begin{subequations}
  \begin{align}
    \mathbf{z}_t^{n,a} = f_{\varphi ^{n,a}}(\boldsymbol{\mathcal{O}}_t^n), \label{socialNavEncoder1} \\
    \mathbf{z}_t^{n,c} = f_{\varphi ^{n,c}}(\boldsymbol{\mathcal{O}}_t^n). \label{socialNavEncoder2}
  \end{align}
\end{subequations}
The network parameters $\phi^n$, $\psi_{1,2}^n$, $\bar{\psi}_{1,2}^n$, $\varphi^{n,a}$, and $\varphi^{n,c}$ are updated using the SAC algorithm \cite{haarnoja2018soft}.

\textit{Remark 2: We employ the SAC algorithm because, in our empirical evaluations, PPO failed to converge on a feasible navigation policy. Furthermore, we utilize only the encoder network; incorporating a decoder within an auto-encoder framework did not provide additional improvements to learning efficiency or navigation performance.}

\textit{Remark 3: We only update the parameters of the CNN encoder together with the critic networks, and then copy the updated parameters from $\varphi^{n,c}$ to $\varphi^{n,a}$. This approach avoids duplicating CNN updates in the actor network, whose gradients are typically much noisier, thereby improving learning stability.}

\subsection{Rewards}

\begin{figure}[!t]
	\centering
	\small
	\includegraphics[width=8.5cm]{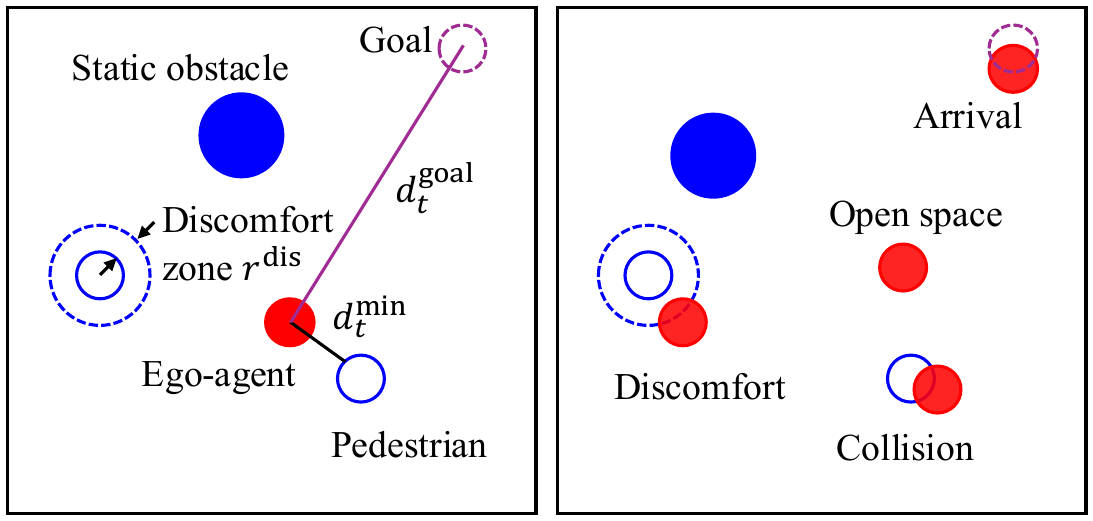}
	\caption{Reward structure and scenarios. The left panel illustrates the geometric representation used for defining the reward function, while the right panel depicts four possible ego-agent scenarios: collision, goal arrival, discomfort, and open space.}
	\label{socialNavReward}
\end{figure}

Since the navigation task prioritizes collision avoidance and goal reaching, the reward function is defined based on four distinct scenarios: goal arrival, collision, discomfort, and open space, as illustrated in Fig. \ref{socialNavReward}:
\begin{equation}
    r_t^n =
        \begin{cases}
        0.5, & d_t ^{\mathrm{goal}} \leq r^{\mathrm{robot}}, \\
        -0.5, & d_t^{\mathrm{min}} \leq r^{\mathrm{robot}}, \\
        r_t^{\mathrm{dis}} + r_t^{\mathrm{goal}}, & r^{\mathrm{robot}} < d_t^{\mathrm{min}} \leq r^{\mathrm{robot}} + r^{\mathrm{dis}},\\
        r_t^{\mathrm{goal}}, & r^{\mathrm{robot}} + r^{\mathrm{dis}} < d_t^{\mathrm{min}},
        \end{cases}
    \label{rewardSocialNav}
\end{equation}
where $r^{\mathrm{robot}}$ denotes the radius of the ego-agent, $d_t^{\mathrm{goal}}$ is the distance between the goal and the ego-agent at time $t$, $d_t^{\mathrm{min}}$ represents the minimum range of the current LiDAR scan, and $r^{\mathrm{dis}}$ defines the annulus width of the discomfort zone, as illustrated in Fig. \ref{socialNavReward}. The discomfort penalty $r_t^{\mathrm{dis}}$ and the goal proximity reward $r_t^{\mathrm{goal}}$ are computed as:
\begin{equation}
    \begin{split}
    r_t^{\mathrm{dis}} &= w^{\mathrm{dis}} \cdot (d_t^{\mathrm{min}} - r^{\mathrm{robot}} - r^{\mathrm{dis}}),\\
    r_t^{\mathrm{goal}} &= w^{\mathrm{goal}} \cdot (d_{t-1} ^{\mathrm{goal}} - d_t ^{\mathrm{goal}}), 
    \end{split}
    \label{rewardSocialNavDisGoal}
\end{equation}
where $w^{\mathrm{dis}}$ and $w^{\mathrm{goal}}$ are the weights for the discomfort penalty and the goal proximity reward, respectively. These terms are designed to further ensure safety and to encourage motion toward the final goal. In this work, we set $r^{\mathrm{robot}} = 0.3~\mathrm{m}$, $r^{\mathrm{dis}} = 0.5~\mathrm{m}$, $w^{\mathrm{dis}} = 0.4$, and $w^{\mathrm{goal}} = 1.4$.

\subsection{Pedestrian Motions and Domain Randomization}
Interactive pedestrian motions are generated using ORCA \cite{van2011reciprocal, alonso2013optimal}, a framework widely adopted in learning-based navigation research \cite{fan2020distributed, chen2019crowd, chen2017decentralized, chen2020relational, jin2020mapless, zhu2024learn}. Unlike standard setups where pedestrians always avoid the robot, we modify the simulation so that pedestrians ignore the ego-agent if their current speed is less than 1.5 times the agent's speed. This modification forces the ego-agent to actively navigate around pedestrians, ensuring the policy learns proactive avoidance rather than relying on the pedestrians' reciprocal collision avoidance strategies.

Since ORCA serves as an ideal multi-agent motion planner for generating pedestrian velocities, we perturb these velocities with uniform noise $\in [-0.15, 0.15]$ m/s to facilitate sim-to-real transfer. Similarly, we apply uniform noise $\in [-0.03, 0.03]$ m to the positions of static obstacles. Because the LiDAR scan construction is derived from these perturbed pedestrian and obstacle positions, no additional domain randomization is applied directly to the LiDAR observations.

\begin{algorithm}
\caption{TRAIN-Nav}
\label{trans_nav}
\begin{algorithmic}[1]
\STATE \textit{Initialize}: $\mathrm{itr} \leftarrow 1$, SAC experience pool $\mathcal{D}$
\STATE \textit{Reset}: ego-agent pose in the world frame, number of static obstacles and pedestrians, positions and sizes of static obstacles, pedestrian positions, $\mathrm{reset}$ $\leftarrow$ \textit{False}, $t \leftarrow 0$
\STATE \textit{Observe}: $\mathbf{a}_{t-1}^n$, $\boldsymbol{g}_t$ and $\boldsymbol{\mathcal{O}}_t^n$
\WHILE{$\mathrm{itr} \leq \mathcal{I}$}
    \STATE Compute $\mathbf{z}_t^{n,a}$ using Eq. (\ref{socialNavEncoder1}) with $\boldsymbol{\mathcal{O}}_t^n$ 
    \STATE Form $\mathbf{s}_{t}^{n,a}$ with $\mathbf{z}_t^{n,a}$, $\mathbf{a}_{t-1}^n$ and $\boldsymbol{g}_t$
    \STATE Sample action $\mathbf{a}_{t}^n$ with $\pi _{\phi ^n}$ and $\mathbf{s}_{t}^{n,a}$
    \STATE Rollout one step with the action $\mathbf{a}_{t}^n$
    \STATE $t \leftarrow t+1$
    \STATE \textit{Observe}
    \STATE Calculate reward $r_t^n$ using Eq. (\ref{rewardSocialNav})
    \IF{arrival \OR collision \OR episode timeout}
        \STATE $\mathrm{reset}$ $\leftarrow$ \textit{True}
    \ENDIF
    \STATE Fill $\mathcal{D}$ with $r_t^n$, $\mathbf{a}_{t-1}^n$, $\boldsymbol{g}_t$, $\boldsymbol{\mathcal{O}}_t^n$ and $\mathrm{reset}$
    \STATE Sample batches from $\mathcal{D}$
    \STATE Update network parameters $\phi ^n$, $\psi _{1,2} ^n$, $\varphi ^{n,a}$ and $\varphi ^{n,c}$ using SAC \cite{haarnoja2018soft}
    \STATE $\mathrm{itr}$ $\leftarrow$ $\mathrm{itr} + 1$
    \IF{$\mathrm{reset}$}
        \STATE \textit{Reset}
        \STATE \textit{Observe}
    \ENDIF
\ENDWHILE
\end{algorithmic}
\end{algorithm}

A summary of the navigation workflow, TRAIN-Nav, is presented in \textbf{Algorithm} \ref{trans_nav}.

\section{Unified Quadrupedal Navigation} \label{unified_quadruped_navigation}
To incorporate quadrupedal dynamics, kinematic constraints, terrain information, and body contact states, we propose a unified architecture that embeds the locomotion policy developed in Section~\ref{quadruped_locomotion} into the navigation strategy introduced in Section~\ref{social_navigation}.

\subsection{Expansion of Navigation}
The unified architecture, illustrated in Fig. \ref{overallFramework}, extends the navigation framework with the following enhancements and modifications:
\begin{itemize}
    \item The simulation employs a high-fidelity quadruped robot model within IsaacSim (see Fig.~\ref{introductionFigure}), replacing the simplified differential kinematics defined in Eq.~(\ref{differentialKinematics}). This configuration enables the framework to fully capture complex quadrupedal dynamics, such as inertia and gait stability, during navigation tasks. Furthermore, the robot is randomly initialized on uneven terrain, as shown in Fig.~\ref{introductionFigure}, ensuring that terrain-specific constraints and proprioceptive feedback are directly integrated into the learning process.

    \item We constrain the lateral velocity $v_y$ to zero because our navigation action space is restricted to linear and angular velocities $(v_x, \omega)$. Our empirical observations indicated that lateral movement frequently caused leg self-collisions, particularly when traversing challenging terrains.

    \item To facilitate the integration, we freeze the locomotion policy pre-trained in Section~\ref{unified_quadruped_navigation}-A, with the lateral velocity command $v_y$ disabled, and interface it directly along with TRAIN-Nav. We freeze the locomotion policy because its control frequency (50 Hz) is an order of magnitude higher than the planning frequency (5 Hz). It is typically challenging to directly backpropagate gradients or simultaneously updating policy parameters across this 10$\times$ temporal discrepancy. Moreover, our experiments revealed that fine-tuning TRAIN-Loco on command distributions sampled from the unified navigation policy led to failure when traversing challenging terrains. The unified navigation policy naturally produces cautious, lower-velocity commands near dynamic agents. Furthermore, the maximum linear velocity is limited to 0.5 m/s for safe navigation, which however constrains its ability to traverse challenging terrain. Fine-tuning on these conservative commands reduces the broad-spectrum dynamic exploration learned by the original terrain-aware locomotion controller.

    \item The unified state space integrates the state representations from both the locomotion controller and the navigation policy, and is defined as follows:
    \begin{equation}
        \mathbf{s}_t^{u,*}=\left[\mathbf{z}_t^{n,*}, \boldsymbol{g}_t, \mathbf{a}_{t-1}^n, \hat{\mathbf{v}}_t, \mathbf{z}_t^l, \mathbf{o}_t^l\right] \in \mathbb{R}^{118},~* \in \{a,c\}.
        \label{unifyStates}
    \end{equation}
    To account for the frequency discrepancy between the navigation and locomotion layers, we update $\hat{\mathbf{v}}_t$, $\mathbf{z}_t^l$, and $\mathbf{o}_t^l$ only after the locomotion policy has executed 10 control cycles. Both $\hat{\mathbf{v}}_t$ and $\mathbf{z}_t^l$ are latent features derived from the previous 10 observations, as defined in Eq.~(\ref{history_observation_locomotion}). This architecture allows us to omit explicit historical locomotion states, relying instead on the latent features from the most recent control step. This expanded state representation effectively integrates terrain awareness and implicit body contact information into the quadrupedal navigation process.
    
    \item To incorporate quadrupedal locomotion safety into the reward function, a penalty of -0.5 is added to Eq.~(\ref{rewardSocialNav}) whenever the robot falls, such as through illegal body contact with the ground. This additional term encourages the policy to prioritize stable locomotion, thereby enhancing the overall stability of quadrupedal navigation with interaction awareness.

    \item To facilitate more stable sim-to-real transfer, we penalize abrupt changes in angular velocity. Consequently, an additional regularization term is incorporated into Eq.~(\ref{rewardSocialNav}):
    \begin{equation}
        r = -0.01 \cdot|w_t^n-w_{t-1}^n|.
        \label{AngularPenalty}
    \end{equation}

    \item The networks are expanded to accommodate the unified state $\mathbf{s}_t^{u,*}$. With the input dimensionality increasing to 118, the input shapes of the expanded networks are updated to (118,) for the actor and (120,) for the critic. Consequently, the network parameters defined in Section~\ref{social_navigation}-C are scaled to match these expanded input dimensions.

    \item To prevent jitter and ensure stable motion at low speeds, the linear and angular velocities are set to zero if the linear velocity commanded by TRAIN-Nav falls below $0.3$ m/s. This thresholding logic results in the following clipped action definition:
    \begin{equation}
        {\mathbf{a}_t^n } =
            \begin{cases}
            \mathbf{0},   & \text{if } v_t^n < 0.3 ~\text{m/s}, \\
            v_t^n, w_t^n, & \text{else}.
            \end{cases}
        \label{ClipAction}
    \end{equation}
\end{itemize}

\textit{Remark 4: We observed that the quadruped robot tended to freeze or become stuck when the commanded linear velocity was below 0.3 m/s, a behavior particularly prevalent on uneven terrain. Additionally, in-place rotation without forward translation occasionally resulted in self-collisions between the robot's legs. To mitigate these issues, we applied a clipping mechanism to the velocity outputs from TRAIN-Nav.}

\subsection{Network Parameters}
The unified networks maintain the same architecture as described in Section~\ref{social_navigation}-C, with the only modification being the input shapes for the actor and critic to accommodate the expanded state space. We denote the updated network parameters as $\phi^u$, $\psi_{1,2}^u$, $\varphi^{u,a}$, and $\varphi^{u,c}$, which are optimized using the SAC algorithm. To accelerate convergence and enhance performance in complex, uneven environments, we initialize the unified policy by warm-starting these parameters with the weights learned from the original navigation networks illustrated in Section \ref{social_navigation}.

The first layer of the actor network in the navigation-only policy consists of weights $\mathbf{w}_1^n \in \mathbb{R}^{54 \times 1024}$ and biases $\mathbf{b}_1^n \in \mathbb{R}^{1024}$. In the unified policy, this layer is expanded to accommodate the high-dimensional state space, with weights $\mathbf{w}_1^u \in \mathbb{R}^{118 \times 1024}$ and biases $\mathbf{b}_1^u \in \mathbb{R}^{1024}$. We initialize $\mathbf{w}_1^u$ and $\mathbf{b}_1^u$ based on $\mathbf{w}_1^n$ and $\mathbf{b}_1^n$ as follows:
\begin{equation}
    \begin{split}
        \mathbf{w}_1^u[:54] &= \mathbf{w}_1^n, ~\mathbf{w}_1^u[54:] = \mathbf{0},\\
        \mathbf{b}_1^u[:54] &= \mathbf{b}_1^n, ~\mathbf{b}_1^u[54:] = \mathbf{0}.\\
    \end{split}
    \label{initiateUnifiedPolciy}
\end{equation}
A similar procedure is applied to the first layer of the critic network. Because the remaining layers in the actor, critic, and encoder networks, as detailed in APPENDIX \ref{navigationNetworkDetail}, share identical dimensions across both the navigation and unified frameworks, we directly transfer the parameters from the pre-trained navigation networks to initialize the unified networks.

\begin{algorithm}
\caption{TRAIN}
\label{trans}
\begin{algorithmic}[1]
\STATE Pre-train locomotion networks with \textbf{Algorithm} \ref{trans_loco}
\STATE Pre-train navigation networks with \textbf{Algorithm} \ref{trans_nav}
\STATE Expand the navigation networks according to the extended states $\mathbf{s}_t^{u,*}$
\STATE Freeze the locomotion networks and embed them into the expanded networks
\STATE Initialize the parameters of the unified networks with the updated parameters from navigation networks
\STATE Update the parameters using the same procedures illustrated in \textbf{Algorithm} \ref{trans_nav}
\end{algorithmic}
\end{algorithm}

Finally, the unified pipeline, TRAIN, is summarized in \textbf{Algorithm}~\ref{trans}.

\section{Simulation Evaluations for Locomotion} \label{simulation_evaluations_locomotion}
\subsection{Benchmarks}
For a comprehensive evaluation, we compared the following SOTA algorithms:

1) \textbf{Privilege} \cite{miki2022learning}: The actor and critic networks have full access to privileged terrain information, representing the upper-bound performance of the policy.

2) \textbf{DreamWaQ} \cite{nahrendra2023dreamwaq}: Our framework builds upon DreamWaQ, utilizing an asymmetric actor-critic DRL architecture.

3) \textbf{SLR} \cite{chen2024slr}: This policy utilizes a symmetric actor-critic DRL architecture, where an encoder extracts latent features shared by both the actor and the critic networks.

4) \textbf{L2T} \cite{wu2025learn}: This method follows a teacher-student paradigm, where the teacher policy first learns quadrupedal locomotion using full access to privileged information. Subsequently, the student policy is trained to clone the teacher's behaviors using only proprioceptive observations available during deployment.

5) \textbf{TAR} \cite{mousa2025teacher}: A similar teacher-student paradigm is proposed. Differently, by aligning representations to a privileged teacher in simulation via contrastive objectives, the student policy learns structured latent spaces.

6) \textbf{TRAIN-Loco}: The proposed method.

\textit{Remark 5: Intuitively, the Privilege method adapts efficiently to difficult environments because its locomotion policy explicitly leverages ground-truth terrain data. Consequently, we employ this method exclusively as an ideal performance reference (upper bound) rather than a direct baseline for comparison with sensor-limited methods.}

\begin{figure}[!t]
	\centering
	\small
	\includegraphics[width=8.5cm]{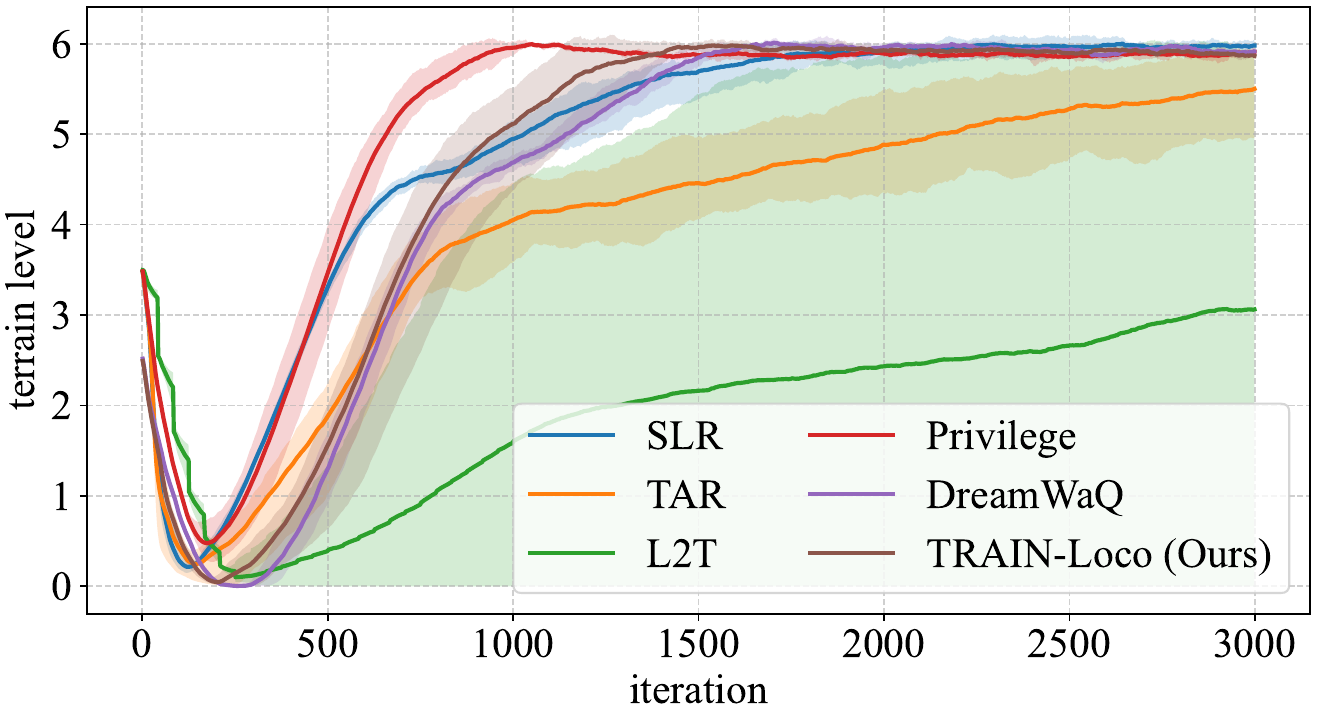}
	\caption{Progression of terrain difficulty levels over training iterations across five independent random seeds. The solid line indicates the mean, while the shaded region denotes the minimum and maximum bounds.. The Privilege method yields an ideal performance reference (upper bound) rather than a direct baseline.}
	\label{locomotionComparison}
\end{figure}

Each of the aforementioned methods was trained using the identical curriculum strategy and domain randomization techniques detailed in Section \ref{quadruped_locomotion}. While DreamWaQ retained its original reward configurations, the other approaches utilized the reward structure defined in TABLE \ref{tableRewardLocomotion}. All training procedures and experiments were conducted using the Unitree Go2 quadruped platform.

Since the focus of our evaluation is terrain-aware locomotion, we utilize the achieved terrain level as the primary performance metric. The training progression is illustrated in Fig. \ref{locomotionComparison} for a maximum of 3,000 iterations. Our framework significantly enhances learning efficiency compared to other methods, particularly teacher-student paradigms (L2T and TAR). Notably, our approach achieves a learning efficiency comparable to the Privilege method, which has full access to ground-truth data. This performance parity suggests that our architecture effectively extracts and utilizes implicit terrain features from proprioceptive history. More quantitative results with respect to velocity tracking performance will be demonstrated in Section \ref{simulation_evaluations_locomotion}-C.

\subsection{Ablations}
To verify the performance improvements of our framework, which builds upon DreamWaQ \cite{nahrendra2023dreamwaq}, we conducted the following ablation studies:

\textbf{TRAIN-Loco-MLP-Go2}: To isolate the impact of our network architecture, we replaced our CNN-based encoder with the MLP architecture from the original DreamWaQ. This variant maintained the Unitree Go2 platform and the fine-tuned reward structure defined in TABLE \ref{tableRewardLocomotion}.

\textbf{Cross-embodiment (Unitree A1)}: We evaluated the portability of our method by replacing the robot platform with the Unitree A1. This set of experiments included:

\begin{itemize}
    \item \textbf{TRAIN-Loco-A1}: Our proposed framework deployed on A1.

    \item \textbf{DreamWaQ-A1}: The original DreamWaQ method adapted for A1.

    \item \textbf{TRAIN-Loco-MLP-A1}: Our framework using DreamWaQ's MLP-based encoder on A1.
\end{itemize}

\begin{figure}[!t]
	\centering
	\small
	\includegraphics[width=8.5cm]{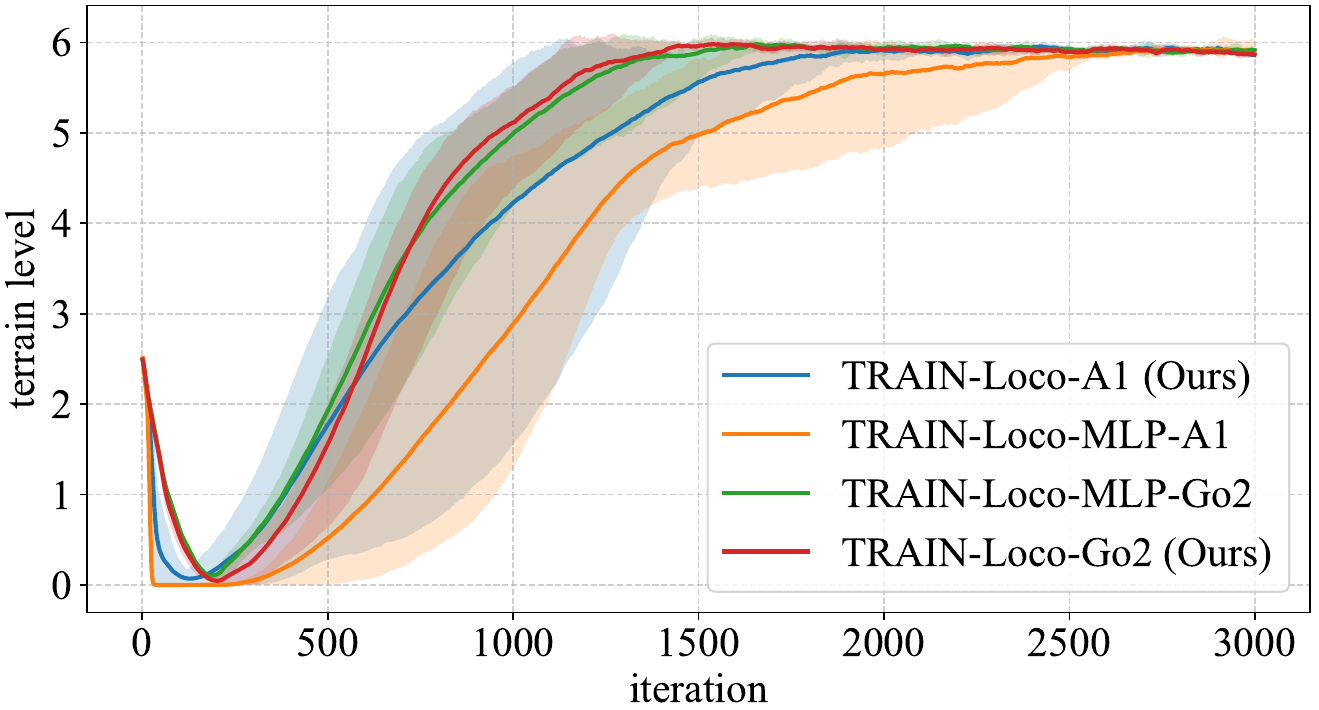}
	\caption{Locomotion ablation results evaluated across five random seeds. DreamWaQ-A1 is omitted because all five runs collapsed (terrain difficulty converged to zero) at the early training stage.}
	\label{locomotionAblation}
\end{figure}

As illustrated in Fig. \ref{locomotionAblation}, the DreamWaQ-A1 variant failed to progress, with its terrain level converging to zero during training. In contrast, our proposed TRAIN-Loco-A1 framework achieved a high terrain level of 6 on both the A1 and Go2 platforms. This performance gap demonstrates that our modifications to the original DreamWaQ architecture significantly enhance cross-platform generalizability.

Furthermore, the results for TRAIN-Loco-MLP-A1 highlight that our fine-tuned reward structure plays a critical role in stabilizing training for the A1 platform. The integration of the CNN-based encoder further accelerated learning efficiency across both quadrupedal platforms, as evidenced by the performance gains in the comparison groups (TRAIN-Loco-MLP-Go2 vs. TRAIN-Loco-Go2 and TRAIN-Loco-MLP-A1 vs. TRAIN-Loco-A1).

\subsection{Command Tracking}

\renewcommand\arraystretch{1.2}
\begin{table}[!t]
	\small
	\centering
	\caption{Command tracking errors for quadrupedal locomotion on uneven terrains.}
	\begin{tabular}{c@{\hspace{6pt}} c@{\hspace{6pt}} c@{\hspace{6pt}} c} 
		\hline
		Method                     & $\Delta v_x$ m/s  & $\Delta v_y$ m/s     & $\Delta w_z$ rad/s    \\
		\hline 
            DreamWaQ-Go2               & 0.12$\pm$0.15     & 0.12$\pm$0.15        & 0.15$\pm$0.15          \\
		    L2T-Go2                    & 0.17$\pm$0.18     & 0.17$\pm$0.17        & 0.28$\pm$0.28           \\
            SLR-Go2                    & 0.13$\pm$0.18     & 0.13$\pm$0.17        & \textbf{0.10}$\pm$\textbf{0.11}           \\
            TAR-Go2                    & 0.45$\pm$0.68     & 0.31$\pm$0.60        & 0.27$\pm$0.34           \\
            TRAIN-Loco-MLP-Go2        & 0.13$\pm$0.15     & 0.12$\pm$0.14        & 0.16$\pm$0.17           \\
            TRAIN-Loco-Go2 (Ours)     & \textbf{0.10}$\pm$\textbf{0.12}     & \textbf{0.11}$\pm$\textbf{0.13}        & 0.13$\pm$0.12           \\
            \hline
            TRAIN-Loco-MLP-A1         & 0.16$\pm$0.20     & 0.14$\pm$0.17        & 0.21$\pm$0.22           \\
            TRAIN-Loco-A1 (Ours)      & \textbf{0.14}$\pm$\textbf{0.15}     & \textbf{0.12}$\pm$\textbf{0.13}        & \textbf{0.15}$\pm$\textbf{0.14}           \\
		\hline
	\end{tabular}
	\label{tableVelocityTrackError}
\end{table}

Since our unified framework relies on high-fidelity velocity tracking during the initial training phase, we further evaluated the command tracking error of our framework against the established benchmarks and ablation models. We randomly distributed 512 robots in simulation and executed 2,000 control steps for each, generating a total of $1,024,000$ (1M+) data points for quantitative analysis. 

We calculated the mean and standard deviation for the tracking errors across linear $xy$ velocities ($\Delta v_x$, $\Delta v_y$) and angular $z$ velocity ($\Delta w_z$). The results, summarized in TABLE \ref{tableVelocityTrackError}, demonstrate that our framework (trained on the Unitree Go2) achieved the highest tracking performance with respect to linear velocities among all non-privileged methods. Although SLR yielded the best tracking for angular velocity, its learning efficiency is notably lower than ours, as illustrated in Fig. \ref{locomotionComparison}. Furthermore, replacing our CNN-based encoder with an MLP-based architecture resulted in performance degradation for both the Go2 and A1 robots, confirming that our encoder can enhance tracking precision.

\begin{figure*}[!t]
	\centering
	\small
	\includegraphics[width=17.5cm]{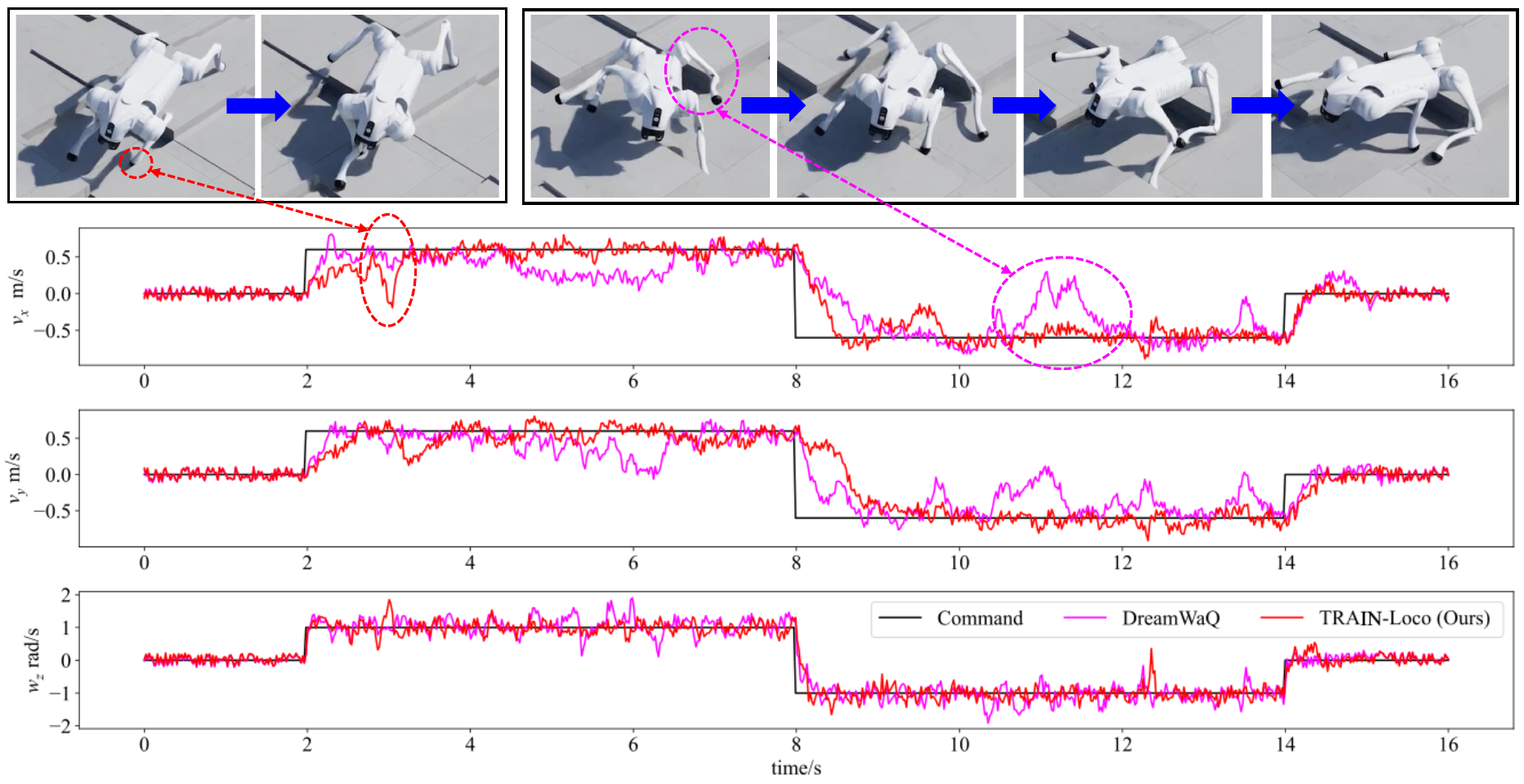}
	\caption{Velocity tracking performance along with simulation scenes. The commanded velocities, $[v_x, v_y, w_z]$, change from $[0.0, 0.0, 0.0]$, to $[0.6, 0.6, 1.0]$, to $[-0.6, -0.6, -1.0]$, and to $[0.0, 0.0, 0.0]$. The top left two scenes correspond to the peak tracking errors for our approach, and the other four illustrate the erroneous tracking for DreamWaQ.}
	\label{realTrackingWithScene}
\end{figure*}

To evaluate real-time performance, we selected a representative velocity tracking sequence and compared our framework against the original DreamWaQ on the Go2 platform, as our locomotion controller builds upon its architecture. The tracking performance and corresponding simulation scenes are illustrated in Fig. \ref{realTrackingWithScene}.

During the evaluation, we observed that DreamWaQ resulted in a lower body clearance; the hind legs tended to splay outward and forward, causing hazardous leg-to-obstacle contact rather than proper foot-end interaction. In contrast, while our approach also encountered occasional tripping from obstacles, the policy demonstrated superior adaptability by quickly repositioning the legs and returning to the commanded velocities. DreamWaQ exhibited a significantly slower recovery after stumbling, further validating the enhanced robustness and velocity tracking performance of our framework.

\section{Simulation Evaluations for Navigation} \label{simulation_evaluations_navigation}

\subsection{Benchmarks}

To provide a rigorous evaluation of our navigation pipeline, we compare it against a diverse set of benchmarks, including classical geometric planners, optimization-based methods, and DRL approaches:

1) \textbf{DWA} \cite{fox2002dynamic}: A classical local planner that samples a discrete velocity space and selects the optimal command based on collision avoidance and goal-reaching criteria.

2) \textbf{ORCA} \cite{van2011reciprocal}: A formal geometric approach that guarantees collision-free motion by assuming reciprocal responsibility among all agents.

3) \textbf{AVOCADO} \cite{martinez2025avocado}: An advancement over ORCA that treats navigation as an adaptive control problem, allowing the robot to adjust to varying levels of cooperation from other agents in real-time.

4) \textbf{N-MPC}: A nonlinear Model Predictive Control strategy that optimizes ego-agent actions over a receding horizon, assuming pedestrians maintain constant velocity.

5) \textbf{NeuPAN} \cite{han2025neupan}: A hybrid method that pre-trains a prediction model to map raw point clouds into a latent distance feature space, which is then utilized by an MPC-based planner for trajectory generation.

6) \textbf{T-MPC} \cite{de2024topology}: A framework that integrates a global planner to identify topologically distinct navigation paths, which serve as constraints for a local MPC solver.

7) \textbf{RGL} \cite{chen2020relational}: A DRL-based policy that utilizes a Relational Graph Learning architecture, operating under the assumption of perfect state information for all pedestrians.

8) \textbf{LNDNL} \cite{zhu2024learn}: A fully end-to-end DRL approach that directly maps raw sensor observations to ego-actions without relying on explicit pedestrian state estimation.

9) \textbf{DRL-VO} \cite{xie2023drl}: A hybrid policy leveraging both pedestrian states and raw observations. To ensure a fair comparison, we integrated the Encoder-integrated SAC structure described in Section \ref{social_navigation} to address original training instabilities.

10) \textbf{HEIGHT} \cite{Liu2026Height}: A hybrid policy leveraging both pedestrian states and raw LiDAR observations. In addition, a heterogeneous spatio-temporal graph is created to model distinct interactions among humans, robots, and obstacles.

11) \textbf{TRAIN-Nav}: The proposed method.

To ensure a fair and consistent evaluation, we standardized the environmental dynamics and agent constraints across all compared methods. We utilized ORCA to simulate deterministic pedestrian motions, intentionally omitting stochastic noise to isolate the performance of the navigation policies. Furthermore, we synchronized the kinematic limits of the ego-agent across all baselines, applying the same linear and angular velocity constraints used in our proposed framework.

\begin{figure}[!t]
	\centering
	\small
	\includegraphics[width=8.5cm]{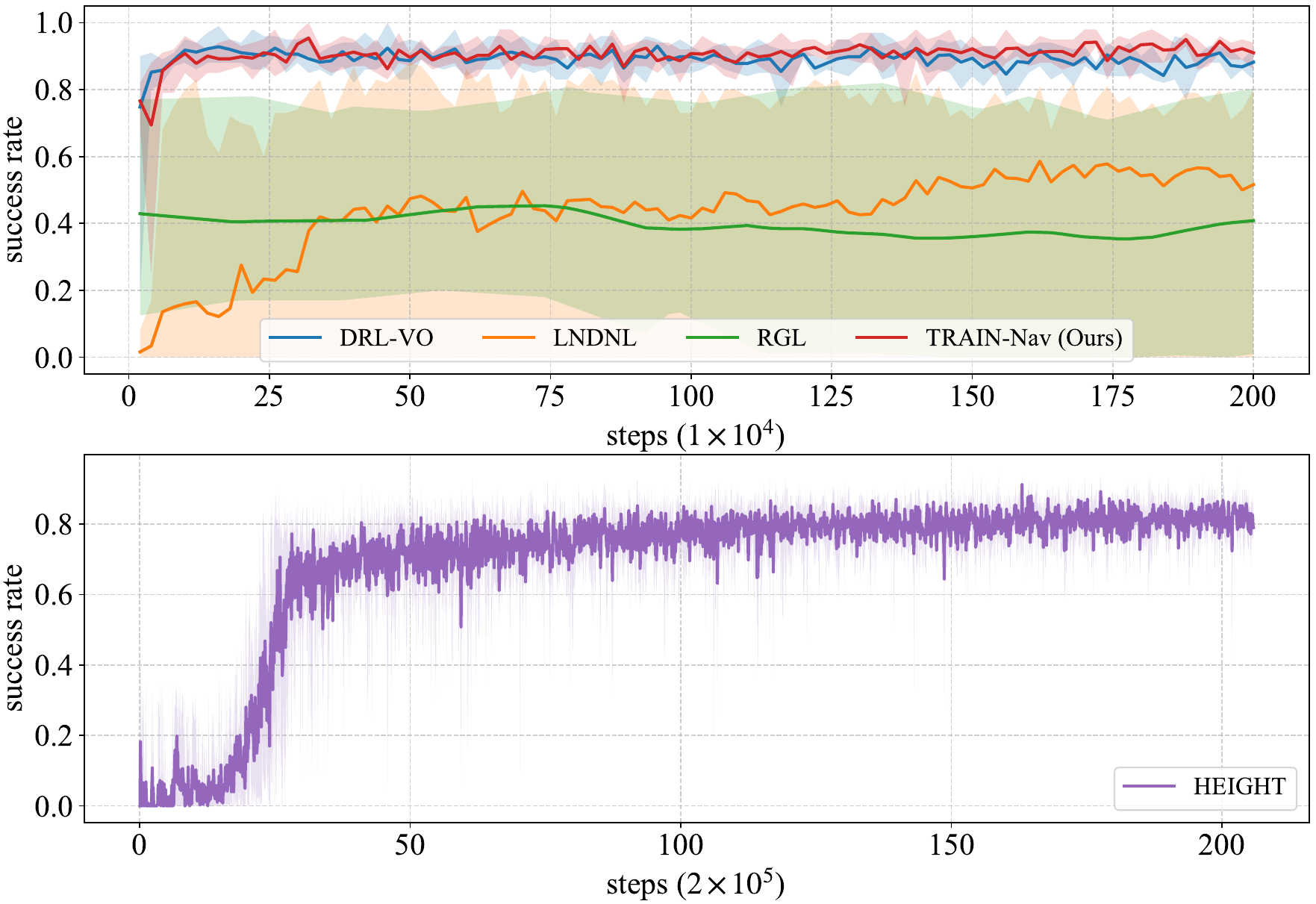}
	\caption{Learning efficiency comparison with five random seeds. The learning curve for HEIGHT is plotted separately in the bottom panel due to its significantly lower learning efficiency (approximately 20 times slower). The remaining learning curves are depicted in the top panel, where DRL-VO achieves an efficiency comparable to our method when utilizing our proposed network architecture.}
	\label{navigationLearningComparison}
\end{figure}

\textbf{Learning performance}. We first evaluated our learning framework against existing DRL-based approaches, with the results illustrated in Fig. \ref{navigationLearningComparison}. We terminated the training of RGL prematurely as its success rate remained significantly lower than other methods; furthermore, its computational overhead was over 15 times that of our approach due to the exhaustive traversal of its discrete action space.

Both RGL and LNDNL converged to suboptimal success rates when subjected to the ego-agent's velocity constraints, suggesting that our navigation framework is more robust under strict kinematic limitations. While DRL-VO achieved a comparable learning efficiency, it did so only after being integrated with our proposed network structure, further validating the effectiveness and stability of our architectural design. HEIGHT converged at a significantly lower speed, and its learning exhibited frequent oscillations, further underscoring the superior learning efficiency and more stable convergence of our navigation framework.

\textbf{Quantitative analysis}. To ensure statistical reliability, we conducted 500 evaluation trials for each method, ensuring that initial environmental configurations were identical across all test cases. Navigation performance was quantified using the following metrics:

\noindent \textbf{Success Rate} (\texttt{SR}): Percentage of trials where the agent reached the goal safely.

\noindent \textbf{Collision Rate} (\texttt{CR}): Percentage of trials resulting in contact with obstacles or pedestrians.

\noindent \textbf{Timeout Rate} (\texttt{TR}): Percentage of trials where the agent failed to reach the goal within the allocated time limit.

\noindent \textbf{Average Navigation Time} (\texttt{NT}): The mean time taken to reach the goal, calculated exclusively for successful trials to avoid skewing data with failed attempts.

\renewcommand\arraystretch{1.2}
\begin{table}[!t]
	\small
	\centering
	\caption{Navigation evaluations with 500 random tests.}
	\begin{tabular}{c c c c c c} 
		\hline
		& Method             & \texttt{SR}      & \texttt{CR}  & \texttt{TR}     & \texttt{NT} \\
		\hline 
		\multirow{3}{*}{Reactive} & DWA                & 0.946            & 0.014            & 0.040           & 18.81$\pm$3.36\\
        & ORCA               & 0.848            & 0.004            & 0.148           & 20.33$\pm$5.59\\
		& AVOCADO            & 0.830            & 0.044            & 0.126           & 18.56$\pm$4.61\\
        \hline
        \multirow{3}{*}{MPC} & N-MPC              & 0.242            & \textbf{0.002}   & 0.756           & 19.89$\pm$3.85\\
        & NeuPAN             & 0.004            & 0.020            & 0.976           & 40.20$\pm$0.85\\
        & T-MPC              & 0.326            & 0.586            & 0.106           & 19.50$\pm$7.31\\
        \hline
        \multirow{3}{*}{DRL} & RGL                & 0.788            & 0.072            & 0.140           & \textbf{18.06}$\pm$3.92\\
        & LNDNL              & 0.926            & 0.022            & 0.052           & 27.88$\pm$6.13\\
        & HEIGHT             & 0.862            & 0.138            & 0.000           & 20.73$\pm$0.89\\
        & DRL-VO             & 0.928            & 0.072            & 0.000           & 19.06$\pm$1.92\\
        \hline
		\multicolumn{2}{c}{TRAIN-Nav (Ours)}  & \textbf{0.972}   & 0.028            & \textbf{0.000}  & 19.26$\pm$1.29\\
		\hline
	\end{tabular}
	\label{tableSocialNavigationComparison}
\end{table}

Compared to the DRL-based baselines, RGL, LNDNL, HEIGHT, and DRL-VO, our framework demonstrated superior performance as evidenced by a higher \texttt{SR}, a lower \texttt{CR}, and a zero \texttt{TR}, as shown in TABLE \ref{tableSocialNavigationComparison}. Although RGL exhibited a shorter \texttt{NT}, our analysis indicates this was due to a reckless policy that prioritized a direct trajectory toward the goal while neglecting necessary collision avoidance. While the original RGL can achieve a comparable \texttt{SR} in unconstrained environments, its performance significantly degrades under our specific testing conditions, highlighting our framework’s superior ability to maintain safety and efficiency under strict action and velocity limitations.

DRL-VO also pre-processes LiDAR scans, utilizing a history of 10 scans where a combination of minimum and average pooling is applied to each. The resulting pooled data is stacked four times to generate a 2D LiDAR map. We observed that relying solely on this LiDAR map as an observation led to a failure in learning. Consequently, DRL-VO requires an object map—recording the positions and velocities of all entities, including static obstacles and moving pedestrians—to capture interactive features. In contrast, our approach relies only on transformed LiDAR scans yet still yields superior navigation performance, demonstrating the effectiveness of our proposed observation space.

In addition to DRL-based methods, we compared our framework against MPC-based trajectory optimization approaches, including N-MPC, T-MPC, and NeuPAN, as summarized in TABLE \ref{tableSocialNavigationComparison}. Both N-MPC and NeuPAN frequently exhibited the ``frozen robot'' problem, as their conservative motion planners failed to find feasible solutions within interactive scenarios. Additionally, the pre-trained prediction model from NeuPAN, which was trained on external datasets, is incompatible with the ORCA-based human motion generator. Conversely, while T-MPC avoided overly conservative behavior, its inherent aggressiveness led to a higher frequency of collisions. A primary factor in the failure of these MPC-based approaches was the application of stricter velocity constraints, which limited the available solution space and necessitated extensive parameter fine-tuning. Similarly, these kinematic constraints degraded the performance of geometric methods such as DWA, ORCA, and AVOCADO, further demonstrating our framework's superior adaptability to highly constrained action spaces.

\begin{figure*}[!t]
  \centering
  \subfloat[TRAIN-Nav (Ours)]{
    \includegraphics[width=0.47\linewidth]{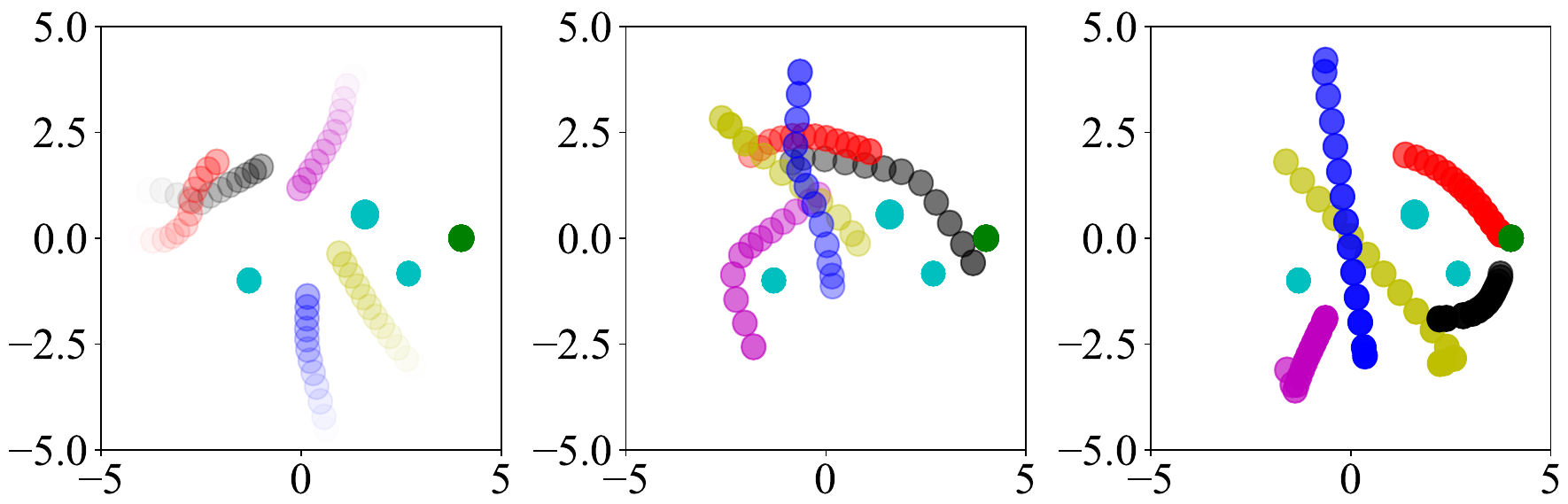}
    \label{trajectory_trans_nav}
  }\hfill
  \subfloat[DRL-VO]{
    \includegraphics[width=0.47\linewidth]{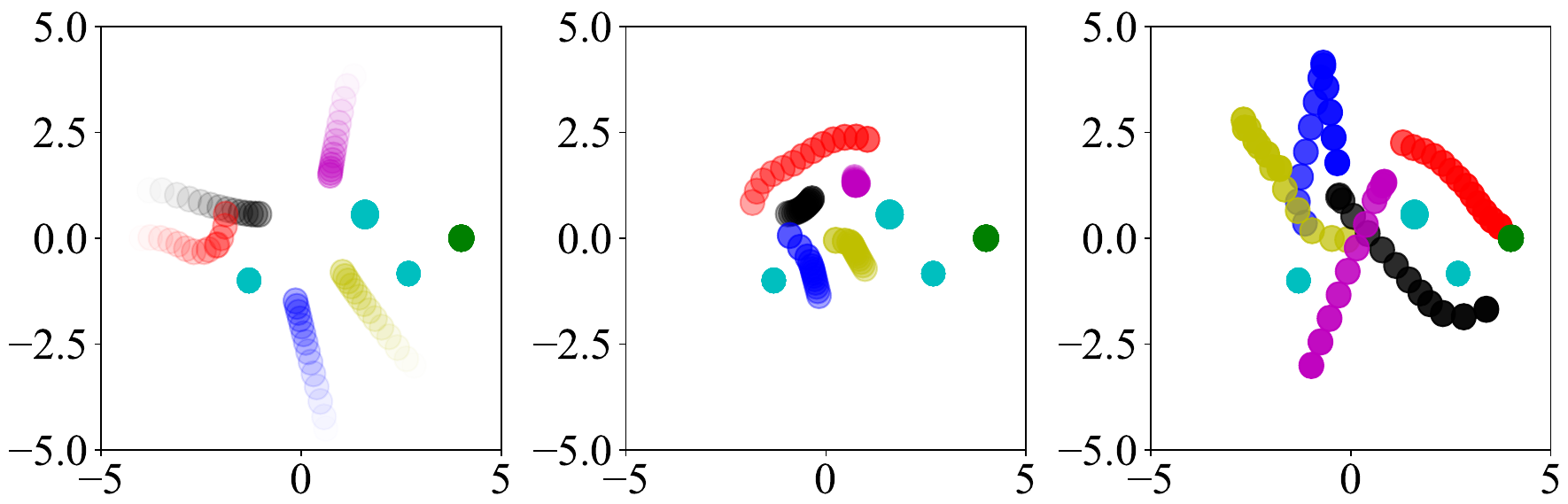}
    \label{trajectory_drl_vo}
  }\\[-1.5ex] 
  \subfloat[HEIGHT]{
    \includegraphics[width=0.47\linewidth]{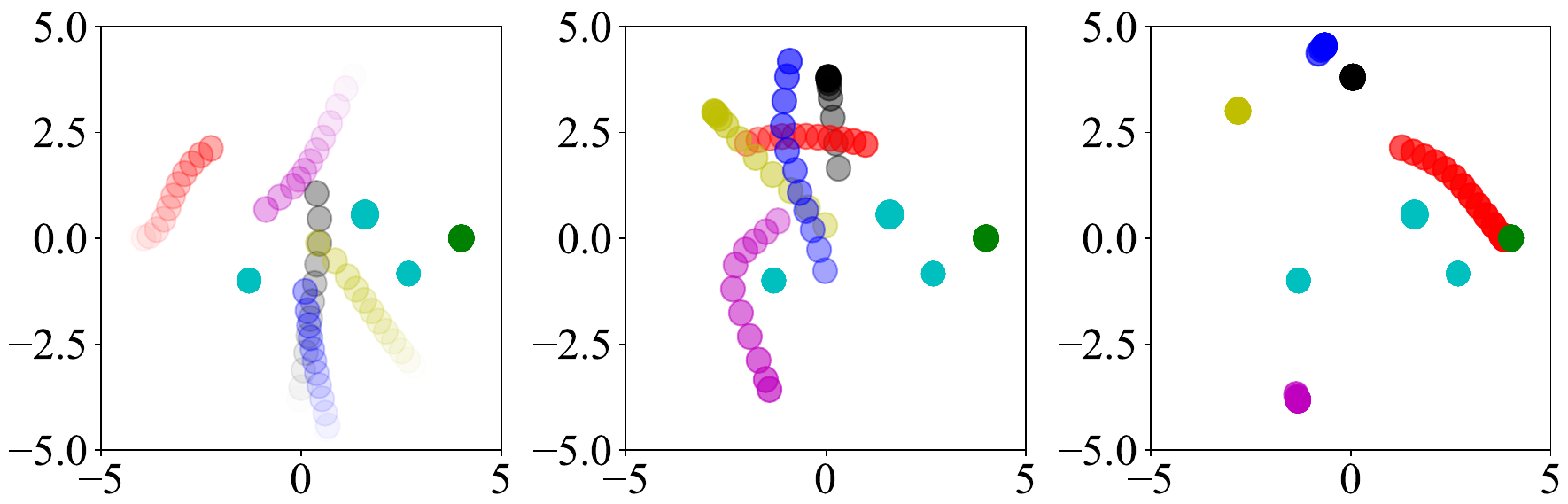}
    \label{trajectory_height}
  }\hfill
  \subfloat[LNDNL]{
    \includegraphics[width=0.47\linewidth]{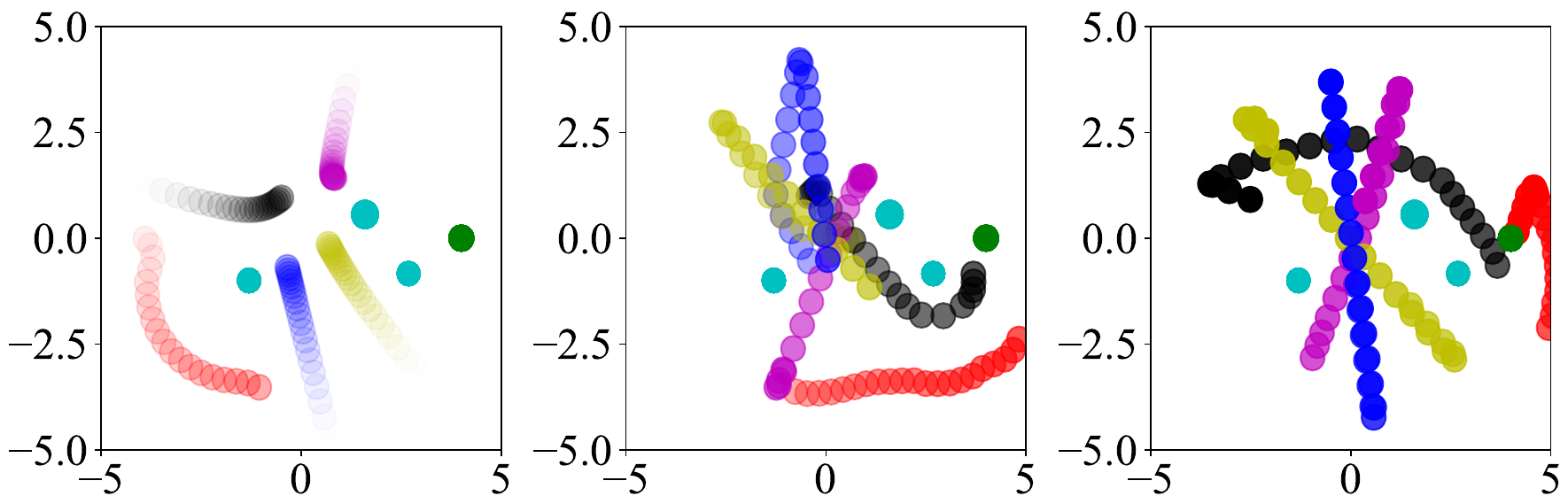}
    \label{trajectory_lndnl}
  }\\[-1.5ex] 
  \subfloat[DWA]{
    \includegraphics[width=0.47\linewidth]{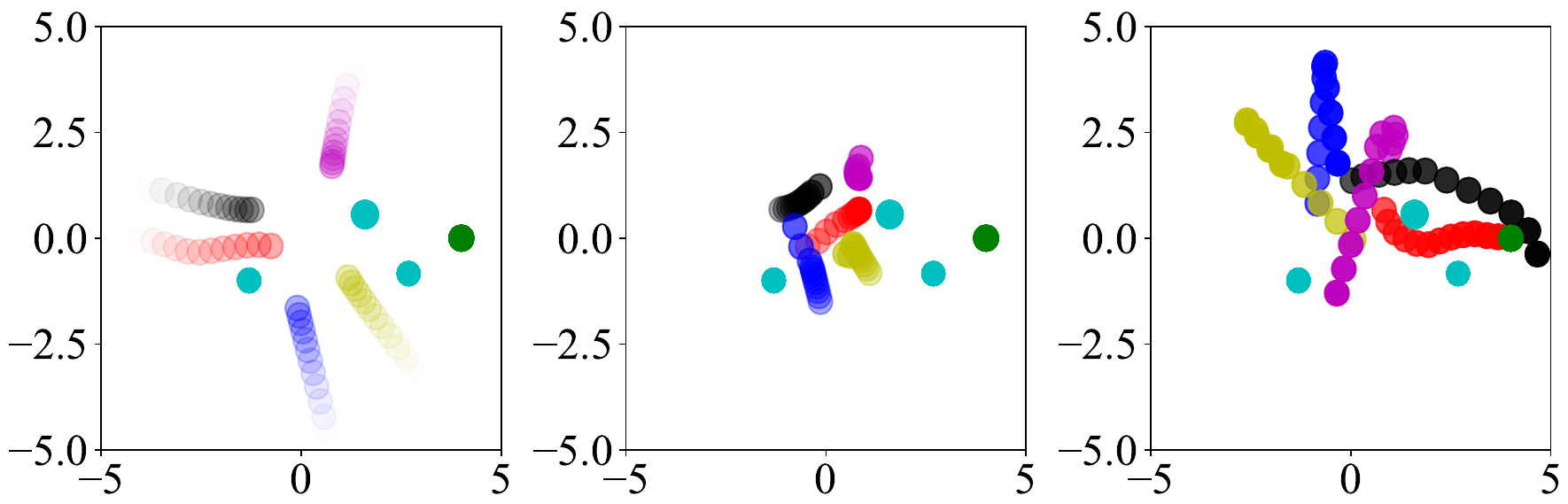}
    \label{trajectory_dwa}
  }\hfill
  \subfloat[ORCA]{
    \includegraphics[width=0.47\linewidth]{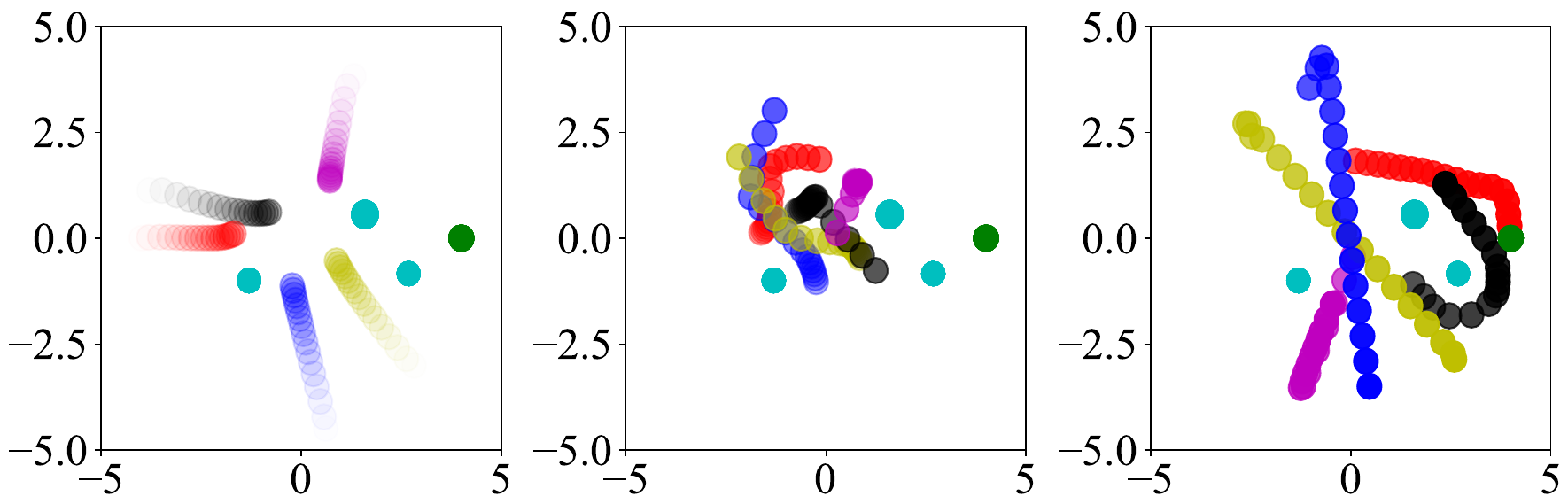}
    \label{trajectory_orca}
  }\\[-1.5ex] 
  \subfloat[T-MPC]{
    \includegraphics[width=0.47\linewidth]{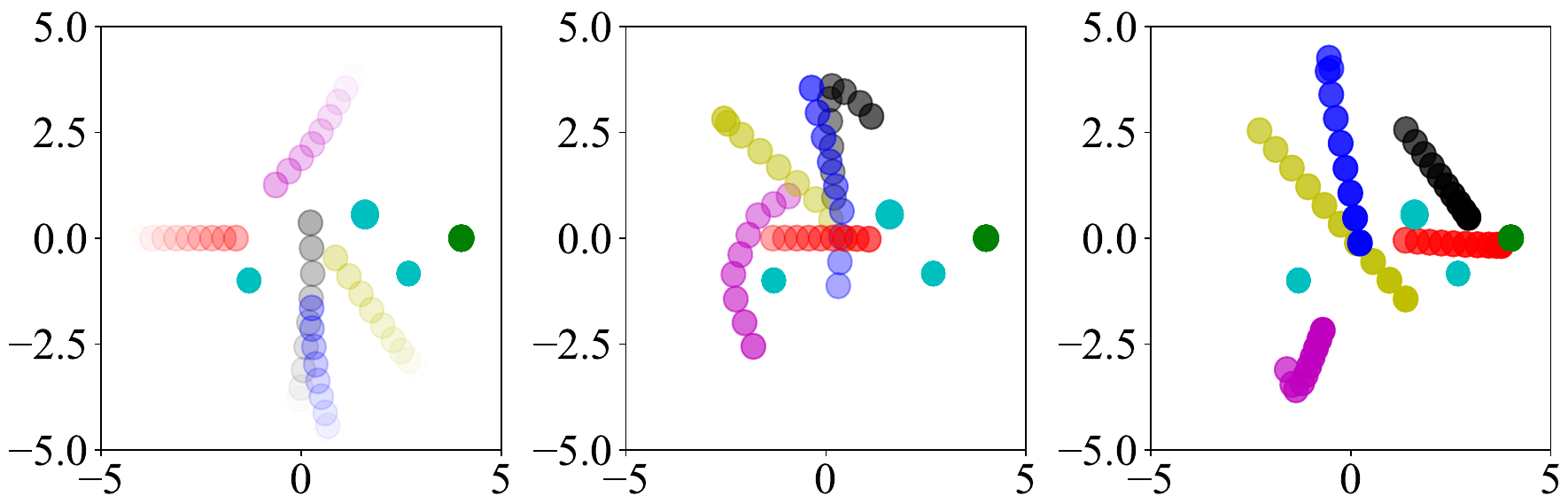}
    \label{trajectory_tmpc}
  }\hfill
  \subfloat[N-MPC]{
    \includegraphics[width=0.47\linewidth]{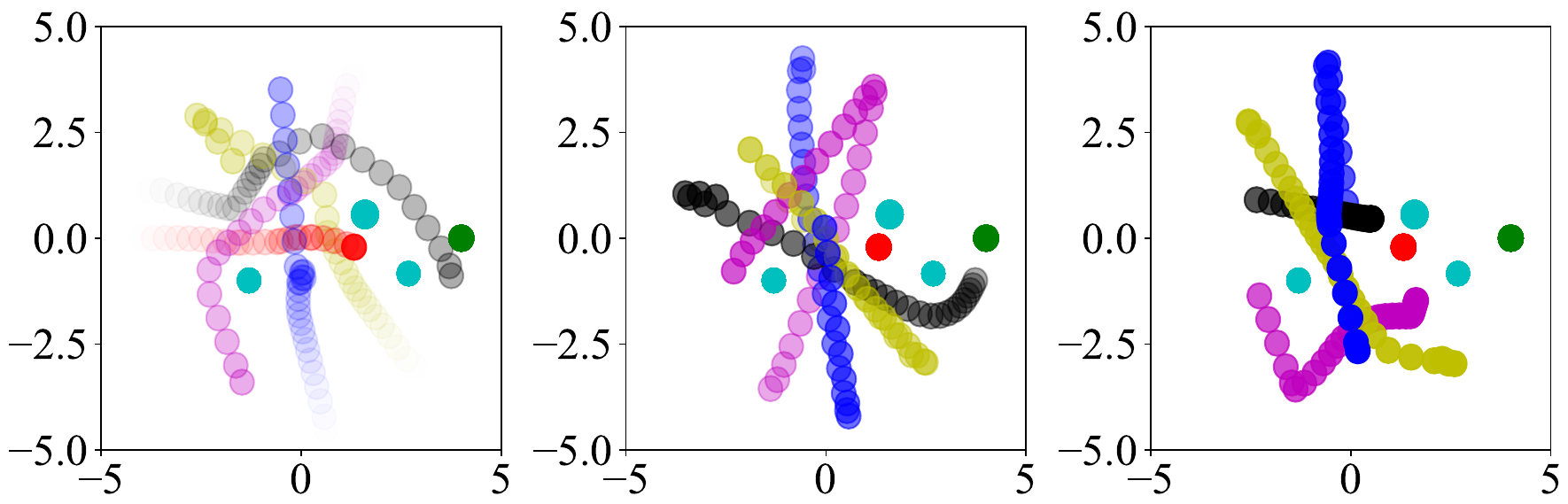}
    \label{trajectory_nmpc}
  }\\[-1.5ex] 
  \subfloat[RGL]{
    \includegraphics[width=0.47\linewidth]{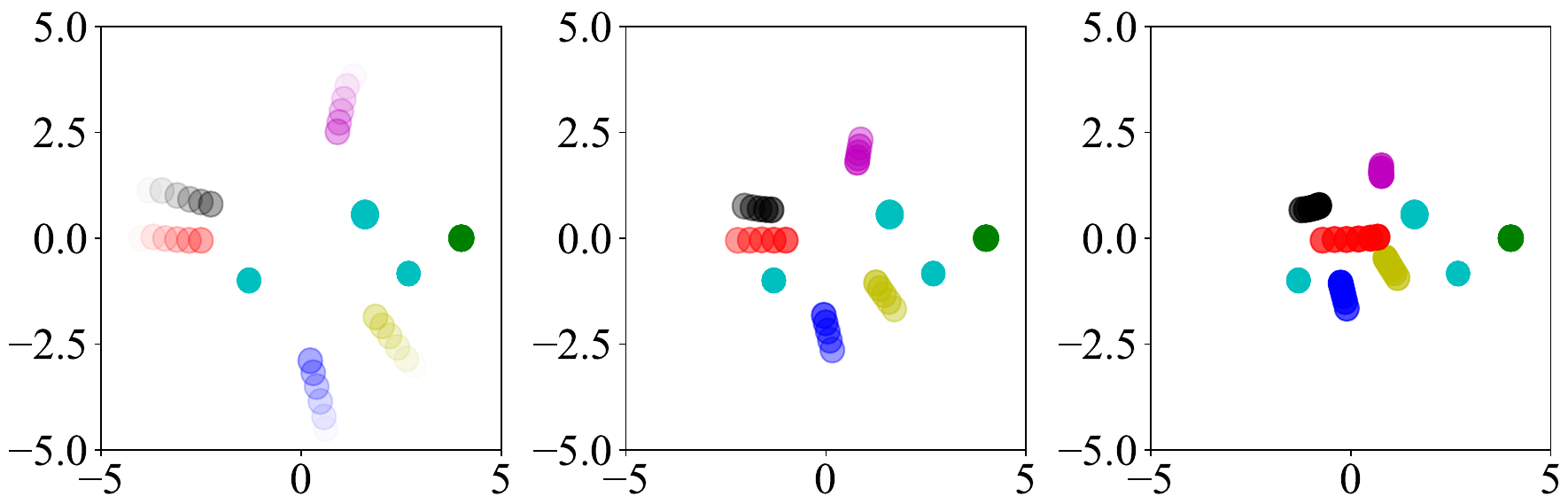}
    \label{trajectory_rgl}
  }\hfill
  \subfloat[AVOCADO]{
    \includegraphics[width=0.47\linewidth]{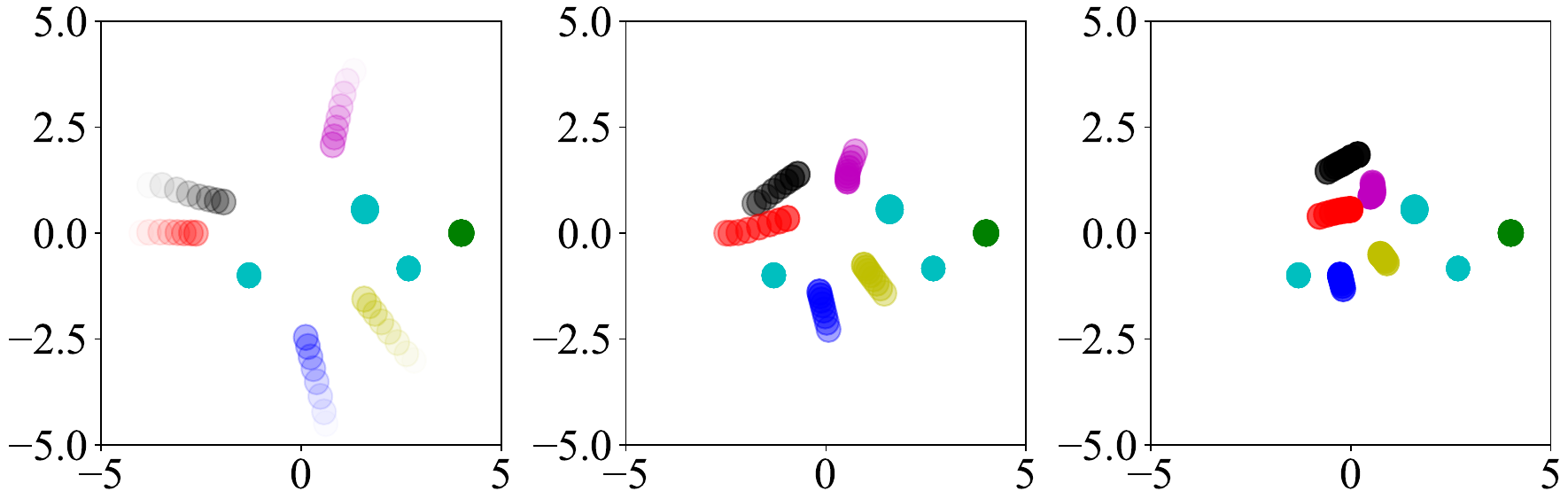}
    \label{trajectory_avocado}
  }\\[-1ex]
  \caption{Trajectories of the ego-agent and pedestrians. The red circle is the ego-agent, the green represents the goal, the cyan circles are static obstacles, and others are pedestrians. As time goes by, the color of the circle changes from light to dark. N-MPC fails due to navigation timeout, while RGL and AVOCADO fail because of collisions.}
  \label{socialNavigationTrajectory}
\end{figure*}

\textbf{Trajectory of ego-agent}. To evaluate the behavioral characteristics of each method, we analyzed a representative scenario featuring three static obstacles and four dynamic pedestrians. The resulting trajectories are visualized in Fig. \ref{socialNavigationTrajectory}, revealing distinct navigation strategies:

\begin{itemize}
    \item \textbf{Aggressive and reckless}: As shown in Fig. \ref{socialNavigationTrajectory}(g), RGL prioritized goal-reaching speed over safety, often failing to maintain adequate clearance for collision avoidance.

    \item \textbf{Conservative and frozen}: N-MPC frequently suffered from ``frozen motion'' due to its conservative optimization constraints (Fig. \ref{socialNavigationTrajectory}(i), right), while LNDNL maintained excessively large safety margins (Fig. \ref{socialNavigationTrajectory}(c)), significantly prolonging travel time.

    \item \textbf{High-risk proximity}: T-MPC was less conservative but navigated dangerously close to obstacles (see Fig. \ref{socialNavigationTrajectory}(f)), leading to the high failure rates noted in TABLE \ref{tableSocialNavigationComparison}. Similarly, DWA often steered the agent into ``discomfort zones'' near obstacles (see Fig. \ref{socialNavigationTrajectory}(d)), increasing collision risks.

    \item \textbf{Interaction bottlenecks}: ORCA trajectories (see Fig. \ref{socialNavigationTrajectory}(e)) exhibited significant deceleration during multi-pedestrian interactions. While AVOCADO (see Fig. \ref{socialNavigationTrajectory}(h)) mitigated this slowdown, it did so at the expense of safety.

    \item \textbf{Optimal balance}: Our framework and the modified DRL-VO produced nearly identical, efficient trajectories, as shown in Fig. \ref{socialNavigationTrajectory}(a) and (b). This convergence further underscores the effectiveness of our network architecture in balancing safety, speed, and constraint satisfaction. Although HEIGHT yielded a similar trajectory, its overall performance was notably inferior, as illustrated in TABLE \ref{tableSocialNavigationComparison}.
\end{itemize}

\subsection{Ablations}
To evaluate the effectiveness of our navigation pipeline, we conducted the following ablation studies:

1) \textbf{TRAIN-Nav-PPO}: We replaced the SAC algorithm with PPO to test the impact of the RL framework. However, the success rate remained at zero throughout the training phase, leading us to exclude this variant from further analysis.

2) \textbf{TRAIN-Nav-CNN}: We enabled CNN updates in both the actor and critic networks to empirically verify the stability and performance claims made in \textit{Remark 3}.

3) \textbf{TRAIN-Nav-Direct}: To assess the impact of our proposed LiDAR scan transformation, we implemented a version that utilizes direct, raw LiDAR scans. This allows us to isolate the benefits of our spatial data processing.

4) \textbf{TRAIN-Nav-Noise}: While our initial comparisons against model-based and DRL-based methods were conducted in noise-free environments for maximum control, we implemented this variant using the domain randomization described in Section \ref{social_navigation}-F. This demonstrates the framework's robustness and its readiness for sim-to-real transfer.

5) \textbf{TRAIN-Nav-Reverse}: For a fair comparison against the baselines in the previous section, backward motion was initially disabled. To further investigate the action-space completeness of TRAIN-Nav, we evaluated a reverse-enabled variant with linear velocity commands ranging over $v_x \in [-0.5, 0.5]\text{ m/s}$.

\begin{table}[!t]
	\small
	\centering
	\caption{TRAIN-Nav: Ablation evaluations with 500 random tests.}
	\begin{tabular}{c c c c } 
		\hline
		Method             & \texttt{SR}      & \texttt{CR}           & \texttt{NT} \\
		\hline 
            TRAIN-Nav-CNN     & 0.892            & 0.108                 & 18.96$\pm$1.40\\
		TRAIN-Nav-Direct  & 0.956            & 0.064                 & 20.66$\pm$2.83\\
            TRAIN-Nav-Noise   & 0.960            & 0.040                 & \textbf{18.75}$\pm$1.48\\
        \hline
		TRAIN-Nav (Ours)  & 0.972   & 0.028        & 19.26$\pm$1.29\\
        TRAIN-Nav-Reverse (Ours)  & \textbf{0.980}   & \textbf{0.020}       & 18.81$\pm$2.29\\
		\hline
	\end{tabular}
	\label{tableSocialNavigationAblation}
\end{table}

Our analysis revealed that while TRAIN-Nav-Direct and TRAIN-Nav-Noise followed training trajectories similar to the original framework, TRAIN-Nav-CNN suffered from significantly degraded learning performance. The quantitative results of these 500 evaluations are summarized in TABLE \ref{tableSocialNavigationAblation}.

Notably, our framework maintained a performance level that outperformed the baselines from TABLE \ref{tableSocialNavigationComparison} even when subjected to noise, underscoring its robustness and potential for sim-to-real transfer. In the TRAIN-Nav-Direct study, the use of raw LiDAR scans without transformation led to a marked decrease in success rate and increase in navigation time; this confirms that our proposed LiDAR transformation more effectively captures the complex dynamics and interactions between the robot and its environment. Even with reverse motion enabled, our approach maintains—and in fact improves—navigation performance without requiring modifications to the architecture or hyperparameters, highlighting the generalizability of the proposed framework. Finally, enabling CNN updates in both the actor and critic networks resulted in a substantial performance drop, empirically validating the architectural advantages detailed in \textit{Remark 3}.

\section{Simulation Evaluations for Unified Policy} \label{simulation_evaluations_unify}

\subsection{Ablations}

To verify the performance improvements provided by our unified policy, we conducted the following ablation studies:

1) \textbf{Nav-Loco}: We decoupled the navigation planner from the locomotion controller. In this setup, the navigation policy first generates a commanded CoM velocity without considering full-body quadrupedal dynamics. These commands are then tracked by our locomotion controller on uneven terrain. This study isolates the benefits of our unified architecture over a traditional hierarchical approach.

2) \textbf{Train-Nav-Loco}: We retrained TRAIN-Nav without the low-level states $\{\hat{\mathbf{v}}_t, \mathbf{z}_t^l, \mathbf{o}_t^l\}$ and omitted the action clipping defined in Eq. (\ref{ClipAction}). Concurrently, the navigation policy was initialized using a warm-start from the original TRAIN-Nav, while the locomotion controller remained frozen.

3) \textbf{TRAIN-Direct}: We retrained the TRAIN framework from scratch using random parameter initialization, omitting both the warm-start phase and action clipping. The target linear velocity $v_t^n$ was sampled continuously from 0.0 to 0.5 m/s.

4) \textbf{TRAIN-Warm-Start}: The network parameters were initialized using the warm-start procedure defined in Eq. (\ref{initiateUnifiedPolciy}), but no action clipping was applied to the output.

5) \textbf{TRAIN-Clip}: We implemented the action clipping strategy defined in Eq. (\ref{ClipAction}) while exempting the network from the warm-start initialization.

6) \textbf{TRAIN-Rotation}: To evaluate stepping-in-place recovery while preventing aggressive yaw rates that may increase the risk of leg self-collision, we formulated a bounded low-speed rotation action mapping:
\begin{equation}
    \mathbf{a}_t^n =
    \begin{cases}
    [0.0, 0.0]^\top, & \text{if } v_t^n < 0.3, ~|\omega_t^n| < 0.3 \\
    [0.0, -0.5]^\top, & \text{if } v_t^n < 0.3, ~\omega_t^n < -0.5 \\
    [0.0, 0.5]^\top, & \text{if } v_t^n < 0.3, ~\omega_t^n > 0.5 \\
    [0.0, \omega_t^n]^\top, & \text{if } v_t^n < 0.3, ~0.3 \le |\omega_t^n| \le 0.5 \\
    [v_t^n, \omega_t^n]^\top, & \text{otherwise}.
    \end{cases}
    \label{rotation_action}
\end{equation}

Furthermore, we implemented DWA as a baseline to generate velocity commands. These commands are subsequently tracked by our locomotion controller, a configuration we refer to as \textbf{TRAIN-Loco-DWA}.

\textit{Remark 6: There are few existing studies addressing quadrupedal navigation in environments that are simultaneously uneven and interactive; thus, we do not implement additional benchmarks. Instead, we leverage a hierarchical pipeline, TRAIN-Loco-DWA, as a baseline. Decoupling the navigation planner from the locomotion controller is a prevailing approach in quadrupedal navigation research.}

\begin{table}[!t]
	\small
	\centering
	\caption{TRAIN: Ablation evaluations and benchmark with 500 random tests.}
	\begin{tabular}{c c c c c} 
		\hline
		Method             & \texttt{SR}      & \texttt{CR}      & \texttt{TR}     & \texttt{NT} \\
		\hline 
		Nav-Loco           & 0.924            & 0.076            & \textbf{0.000}           & 19.62$\pm$2.20\\
        Train-Nav-Loco     & 0.930            & 0.066            & 0.004           & 19.54$\pm$2.26\\
        TRAIN-Direct       & 0.924            & 0.076            & \textbf{0.000}           & 19.60$\pm$2.15\\
        TRAIN-Warm-Start   & 0.954            & 0.042            & 0.004           & 20.39$\pm$2.81\\
        TRAIN-Clip         & 0.916            & 0.076            & 0.008           & 21.28$\pm$3.38\\
        TRAIN-Loco-DWA     & 0.894            & 0.078            & 0.028           & \textbf{18.90}$\pm$3.23\\
        TRAIN-Rotation (Ours) & 0.968 & 0.030 & 0.002 & 19.60$\pm$2.62 \\  
		TRAIN (Ours)       & \textbf{0.976}   & \textbf{0.024}   & \textbf{0.000}           & 20.57$\pm$3.68\\
		\hline
	\end{tabular}
	\label{tableUnifiedPolicyAblation}
\end{table}

\begin{figure}[!t]
  \centering
  \subfloat[Head-on collision]{
    \includegraphics[width=\linewidth]{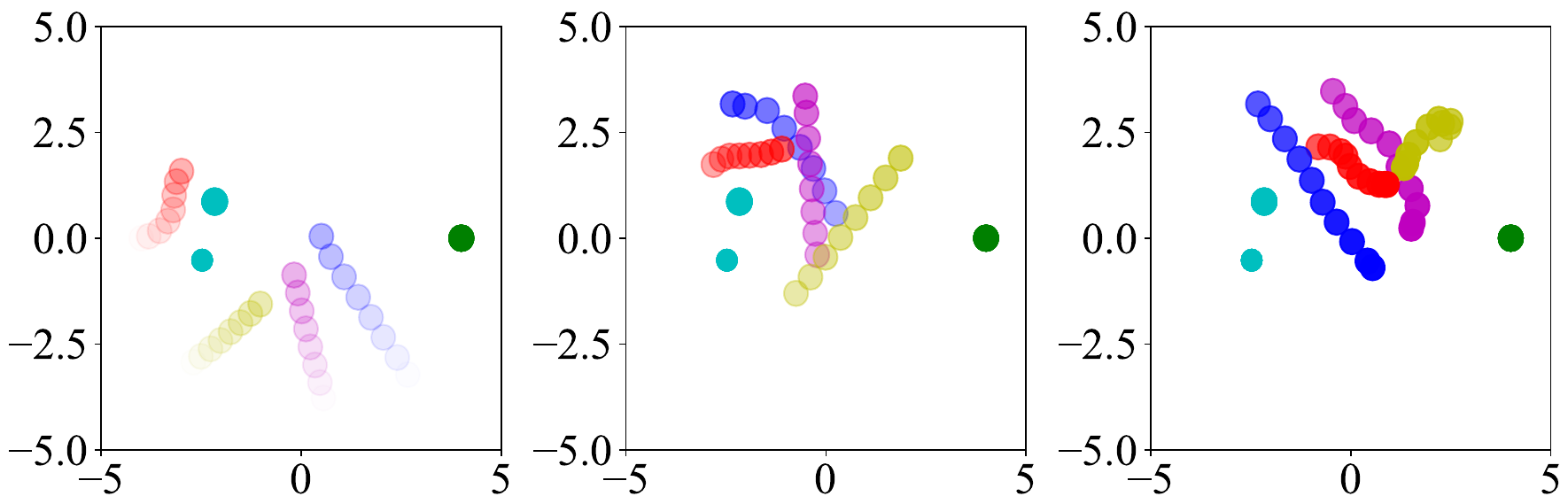}
    \label{bothCollision}
  }\\[-0.3ex]
  \subfloat[Pedestrian rear-end collision]{
    \includegraphics[width=\linewidth]{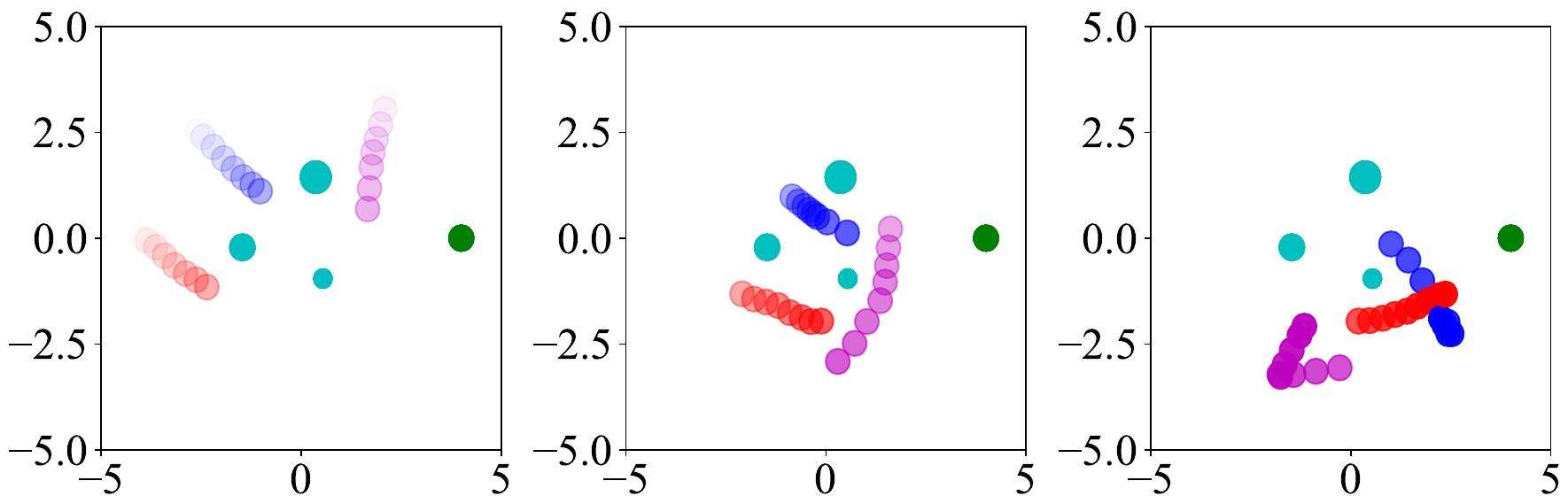}
    \label{pedestrianCollision}
  }\\[-0.3ex]
  \subfloat[Robot rear-end collision]{
    \includegraphics[width=\linewidth]{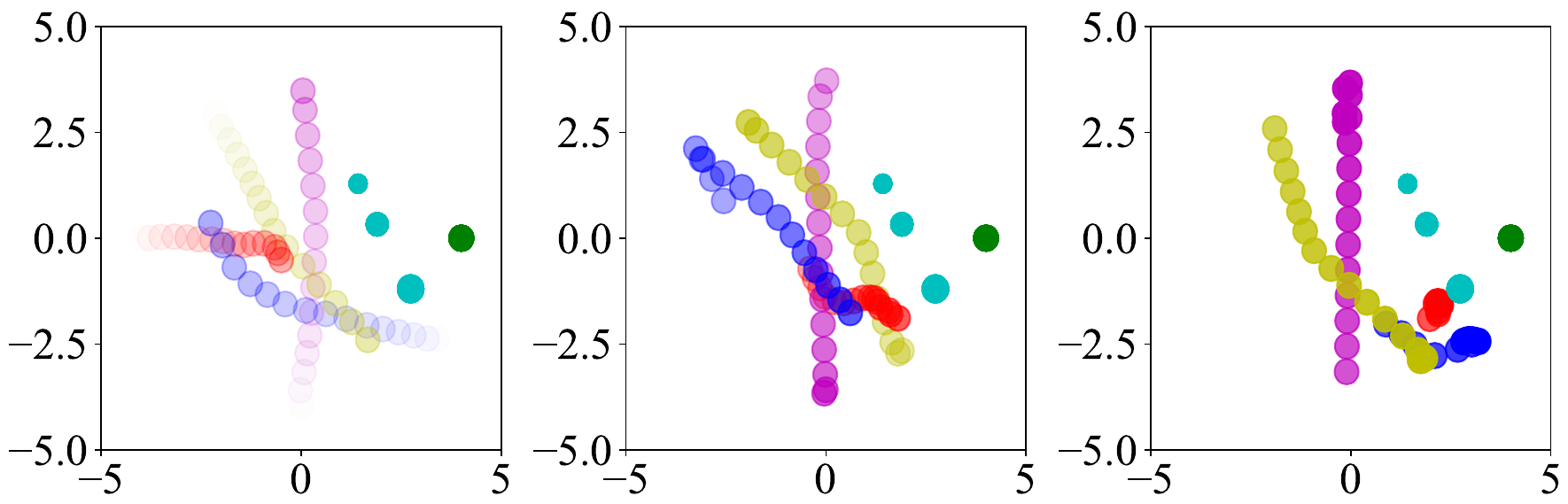}
    \label{robotCollision}
  }\\[-0.3ex]
  \subfloat[Simulation scenes for robot collision]{
    \includegraphics[width=\linewidth]{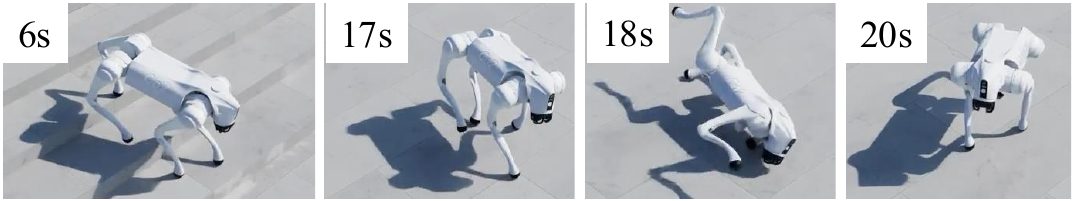}
    \label{simulationScenes}
  }\\[-0.25ex]
  \caption{Trajectories and simulation environments for representative failure cases. (a) Top row: A head-on collision between the robot and a dynamic pedestrian. (b) Middle row: A scenario where a pedestrian rear-ends the robot due to reactivity differences. (c) Third row: Trajectory analysis of a collision with a static obstacle. (d) Bottom row: Corresponding 3D simulation snapshots for the obstacle collision cases presented in (c).}
  \label{transFailures}
\end{figure}

The ablation studies yielded similar learning curves; therefore, we omit a dedicated plot similar to Fig. \ref{navigationLearningComparison}. Instead, we present the final quantitative evaluation across 500 randomized trials, as summarized in TABLE \ref{tableUnifiedPolicyAblation}. Our results indicate that the traditional decoupled pipeline, Nav-Loco, suffers from a reduced success rate due to the gap between high-level planning and low-level execution. Retraining the unified policy from scratch (TRAIN-Direct) yielded performance comparable to the decoupled baseline, suggesting that naive learning alone is insufficient without proper guidance.

However, applying warm-start initialization increased the success rate by 3$\%$, confirming that pre-training with TRAIN-Nav effectively bootstraps quadrupedal navigation on uneven terrains. TRAIN's learning framework is SAC, which leverages an experience replay buffer initialized with transitions collected by the pretrained Stage 1 policy. Because the warm-started policy begins with high initial capability, early exploration populates the buffer with high-reward demonstrations, including successful goal-reaching and effective collision avoidance. Anchoring online exploration with high-quality demonstration data stabilizes early updates and reduces the risk of policy degradation \cite{nair2020awac}. Experimentally, after 20,000 network updates, the warm-started policy achieved a success rate of 0.932, compared with 0.376 for the policy trained from scratch. Furthermore, TRAIN-Warm-Start eventually achieved a superior success rate compared to Train-Nav-Loco (0.954 vs. 0.930), suggesting that the incorporation of low-level states enhances quadrupedal navigation across uneven terrains.

Enabling controlled low-speed rotation substantially expands the low-speed maneuverability of TRAIN. Although operation near zero forward velocity introduces additional gait instability on severe terrain and results in a modest decrease in success rate relative to the high-momentum configuration, the experiment demonstrates that the framework can be extended to support low-speed recovery behaviors. While individual action clipping slightly degraded performance in simulation, it remains a vital component for ensuring safe sim-to-real transfer. Notably, the synergy between warm-start initialization and action clipping significantly boosted the success rate to 0.976, surpassing even the standalone TRAIN-Nav. This peak performance highlights our architecture's exceptional adaptability to the discrete and highly constrained action spaces inherent in legged robot navigation.

The comparison with DWA further validates the enhanced navigation performance provided by our unified pipeline. As shown in TABLE \ref{tableSocialNavigationComparison}, directly deploying a high-level motion planner on a quadrupedal robot in uneven terrain significantly degrades the success rate. The robot is prone to becoming stuck, as evidenced by the increased timeout rate, because the high-level motion planner fails to account for the ``freezing'' phenomenon of quadrupedal locomotion in uneven terrain. Moreover, we observed that the robot tended to fall due to the generation of kinematically unsafe velocity commands. This underscores the necessity of a unified policy, particularly for quadrupedal navigation in interactive and non-planar environments.

\subsection{Failure Analysis}

\begin{figure*}[!t]
	\centering
	\small
	\includegraphics[width=17.5cm]{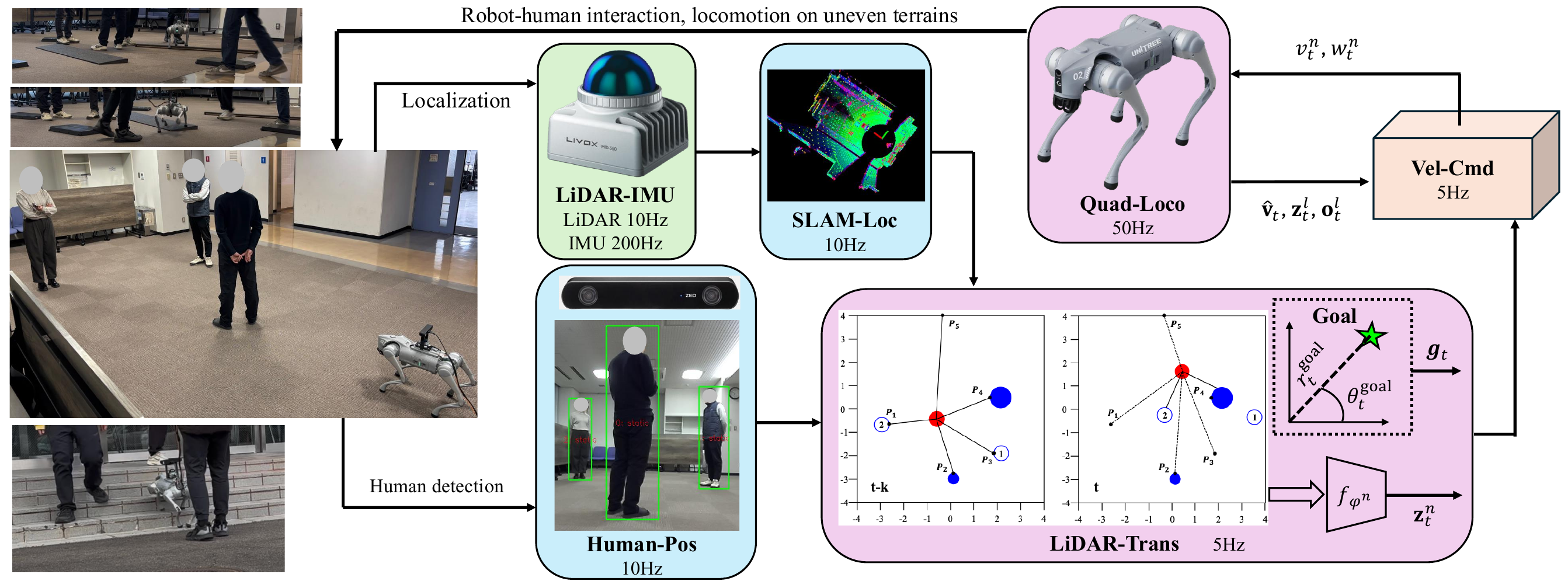}
	\caption{Overview of the hardware deployment pipeline. Visual data for pedestrian tracking is acquired via the onboard camera, while localized navigation is achieved through LiDAR-IMU fusion. These data streams are processed on an external PC to execute the real-time perception-to-action mapping.}
	\label{hardwareWorkflow}
\end{figure*}

Despite the unified policy achieving a high success rate (0.976), infrequent failures were still recorded. As illustrated in Fig. \ref{transFailures}, we categorized the observed collisions into three distinct failure modes: head-on collisions, pedestrian rear-end collisions, and robot rear-end collisions. Out of 500 trials reported in TABLE \ref{tableUnifiedPolicyAblation}, 12 collisions occurred: 3 head-on, 3 robot rear-end, and 6 pedestrian rear-end. Failure analysis is given as follows:

\textbf{Pedestrian dynamics}: Because the pedestrian motion generator is highly reactive while the robot is governed by strict velocity constraints, pedestrians are statistically more likely to rear-end the robot.

\textbf{Mutual avoidance errors}: Despite reciprocal avoidance efforts from both parties, head-on collisions occasionally persist. A potential mitigation strategy involves adopting more conservative behaviors, such as a ``temporary stop'' command upon predicting a high-risk interaction.

\textbf{Terrain and maneuverability}: We observed occasional robot drift caused by challenging terrain, leading to rear-end collisions with pedestrians. Furthermore, the robot occasionally became ``stuck'' near static obstacles. While rotating in place offers a recovery path, it can induce locomotion instability and inadvertently trigger rear-end contact.

To address these limitations, we will focus on:

\textbf{Enhanced rotational stability}: Improving the stability of high-torque rotational maneuvers for quadruped robots on uneven terrain.

\textbf{Recovery behaviors}: Enabling safe backward locomotion to recover from stalled or ``frozen'' states without compromising full-body balance.

\subsection{Zero-Shot Generalization to Unseen Environments}

\begin{figure}[!t]
	\centering
	\small
	\includegraphics[width=6.5cm]{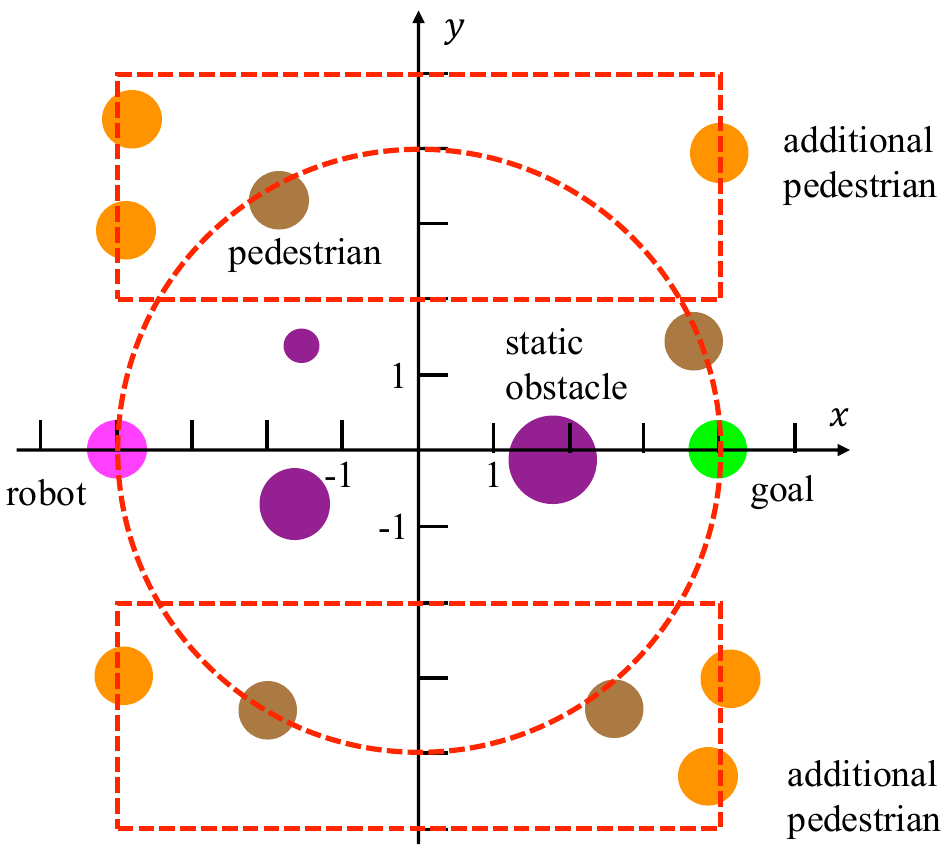}
	\caption{Unseen pedestrian environments. The robot starts at (-4, 0) m with a goal position of (4, 0) m. Static obstacles of varying sizes are randomly distributed along the robot’s path. The initial group of pedestrians is distributed in a circular formation, while the upper and lower rectangles define the movement zones for additional pedestrians.}
	\label{unseenEnvironments}
\end{figure}

In our training environments, the maximum number of pedestrians is four, distributed along a circle with a radius of $4$ m. To evaluate the zero-shot generalization to unseen environments, we increase the maximum pedestrian count to ten. The distribution of these additional pedestrians is illustrated in Fig. \ref{unseenEnvironments}. Specifically, we define two symmetric rectangular zones where additional pedestrians are randomly initialized on the left and right sides. To avoid overly complex interactions or excessive crowding in the center, their goals are positioned on the opposite side of their respective zones. The pedestrians move back and forth between these boundaries to simulate flow environments that were not encountered during training. Moreover, the additional pedestrians also interact with the robot and the original pedestrians. Across 500 test trials, the success rate decreased slightly from $0.976$ to $0.926$, while the average navigation time increased from $19.26$ s to $20.53$ s. Despite increasing the pedestrian density by 1.5 times, the navigation performance exhibited only marginal degradation, further demonstrating the robust zero-shot generalization of our framework to unseen scenarios.

\section{Hardware Implementations} \label{hardware_implementations}
\subsection{Quadrupedal Navigation System}

The hardware configuration and computational workflow for our deployment are illustrated in Fig. \ref{hardwareWorkflow}. We utilize a Unitree Go2 quadrupedal robot equipped with a Livox MID-360 LiDAR and a ZED2 stereo camera. The software architecture is executed on an external laptop (Intel i7-10750H CPU, NVIDIA GTX 1650 Ti GPU) and is comprised of the following functional modules:

\begin{itemize}
    
    \item \textbf{LiDAR-IMU}: Manages raw data acquisition from the LiDAR and IMU sensors, ensuring precise time synchronization before data publishing.

    \item \textbf{SLAM-Loc}: Subscribes to synchronized LiDAR and IMU streams and employs the FAST-LIO algorithm \cite{xu2021fast} to provide high-frequency global localization.
    
    \item \textbf{Human-Pos}: Processes stereo camera data via the ZED2 SDK to detect pedestrians and estimate their spatial coordinates within the camera frame.

    \item \textbf{LiDAR-Trans}: Integrates global localization and human position data to reconstruct and transform historical LiDAR scans. It computes the latent features $\mathbf{z}_t^n$ and determines the goal position $\boldsymbol{g}_t$ relative to the robot frame.

    \item \textbf{Quad-Loco}: Executes the TRAIN-Loco controller to handle full-body locomotion. It subscribes to velocity commands from the high-level navigation planner and publishes the estimated velocity $\hat{\mathbf{v}}_t$, latent locomotion features $\mathbf{z}_t^l$, and proprioceptive observations $\mathbf{o}_t^l$.

    \item \textbf{Vel-Cmd}: Acts as the high-level decision maker, subscribing to all state features, $[\mathbf{z}_t^n, \boldsymbol{g}_t, \hat{\mathbf{v}}_t, \mathbf{z}_t^l, \mathbf{o}_t^l]$, to generate the final velocity commands.

\end{itemize}

To manage inter-module communication and data synchronization, we utilize ROS Noetic with a multi-rate execution strategy: the IMU publishes at 200Hz, Quad-Loco operates at 50Hz, SLAM-Loc and Human-Pos are synchronized at 10Hz, and the high-level LiDAR-Trans and Vel-Cmd modules run at 5Hz. To eliminate computational resource competition, we implemented a dedicated CPU affinity strategy across 12 cores, allocating 4 cores to the computationally intensive LiDAR-Trans module and the remainder distributed among sensing and control tasks. Under full system load, the execution latencies for SLAM-Loc, Human-Pos, LiDAR-Trans, Quad-Loco, and Vel-Cmd were recorded at 2.5, 28.5, 7.6, 1.7, and 0.8 ms, respectively. These results confirm the real-time feasibility of our framework, as the critical control and planning modules operate well within their respective timing budgets.

\subsection{Real-world Implementations}

\begin{table}[!t]
	\small
	\centering
	\caption{Hardware implementations.}
	\begin{tabular}{c c c c c} 
		\hline
		  Scenarios           & flat      & uneven      & slopes     & stairs \\
		\hline   
		  Success/Failure     & 3/0       & 4/1         & 3/0        & 6/4 \\
		\hline
	\end{tabular}
	\label{hardwareResults}
\end{table}

We evaluated our policy across a diverse range of environments, including indoor flat and uneven surfaces, outdoor cement and grass slopes, and stairs with a 12 cm step height. The quantitative success rates are summarized in TABLE \ref{hardwareResults}, while the supplementary videos provide a qualitative demonstration of the system's real-world performance.

The majority of failures were observed during stair traversal, indicating that the simultaneous demand for collision avoidance and stair climbing remains a significant sim-to-real challenge. Furthermore, a recurring challenge observed across all experimental scenarios was localization drift, primarily induced by high-frequency structural vibrations and the presence of dynamic environmental features. These factors significantly degraded real-world performance; consequently, our future research will prioritize the enhancement of localization stability through robust state estimation techniques.

\begin{figure}[!t]
	\centering
	\small
	\includegraphics[width=8.5cm]{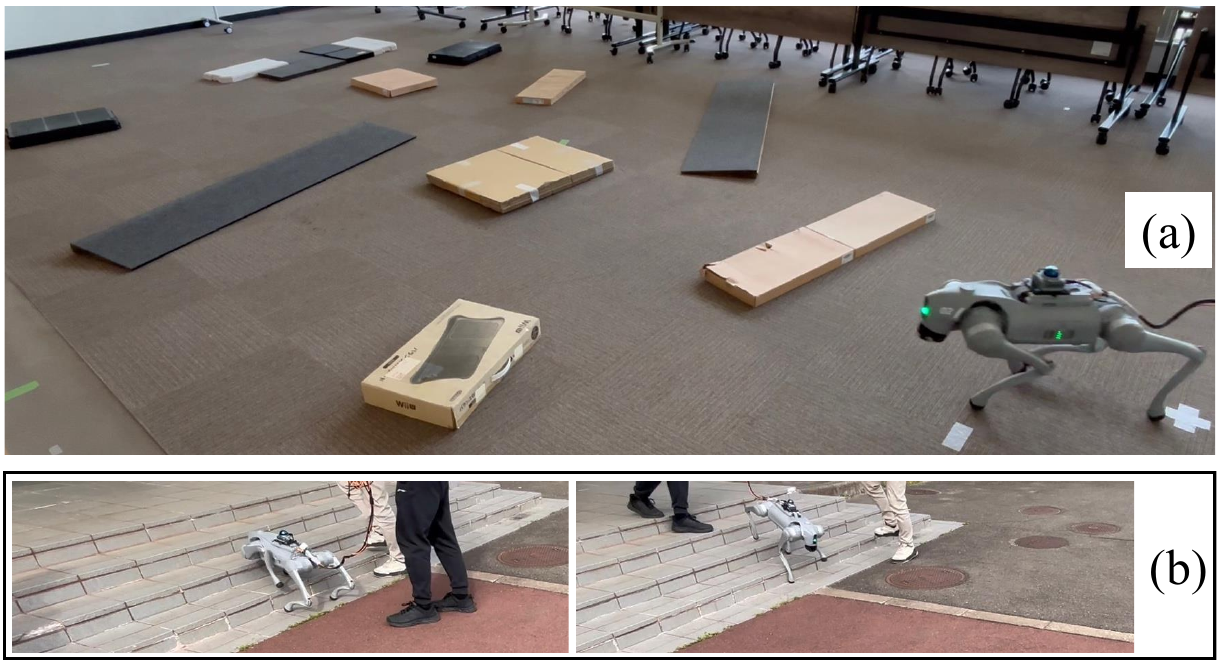}
	\caption{Real-world terrain configurations. The top panel displays the indoor rough terrain, while the bottom panels illustrate the stair scenarios.}
	\label{statisticsScen}
\end{figure}

\begin{table}[!t]
	\small
	\centering
	\caption{Hardware implementations with simulated pedestrians and obstacles.}
	\begin{tabular}{c c c c c c} 
		\hline
		  Method         & \texttt{SR}          & \texttt{NT}                     & Fall          & Collision    & Timeout\\
		\hline   
		  DWA            & 0.82        & \textbf{16.40}         & 3/50             & 2/50            & 4/50  \\
        TRAIN (ours)  & \textbf{0.92}        & 19.15         & \textbf{3}/50    & \textbf{1}/50            & \textbf{0}/50  \\
		\hline
	\end{tabular}
	\label{statisticsIndoor}
\end{table}

To mitigate localization drift caused by highly dynamic environments, we replaced real pedestrians with simulated objects, as detailed in Section \ref{simulation_evaluations_navigation}. The terrain configuration is depicted in Fig. \ref{statisticsScen}-(a). The robot traverses from coordinate (-3.5, 0.0) m to (3.5, 0.0) m. To demonstrate the superiority of our approach, we implemented the DWA as a baseline. Fifty trials were conducted for each method, with the quantitative results summarized in TABLE \ref{statisticsIndoor}. Although DWA had access to pedestrian velocity, its success rate was lower than ours because it failed to account for quadrupedal locomotion constraints and terrain information during motion planning. Moreover, the inference time for DWA was 30 ms, significantly longer than our approach (0.8 ms), further validating the superior real-time performance of our method.

\begin{table}[!t]
	\small
	\centering
	\caption{Hardware implementations with simulated pedestrians and obstacles on stairs.}
	\begin{tabular}{c c c c c} 
		\hline
		               & Success  & Collision   & Fall     & \texttt{NT} \\
		\hline   
		Descending     & 9        & 1           & 0        & 17.38 \\
        Ascending      & 8        & 1           & 1        & 22.65 \\
		\hline
	\end{tabular}
	\label{hardwareResultsStairs}
\end{table}

To obtain statistical results for stair environments, as depicted in Fig. \ref{statisticsScen}-(b), we conducted an additional 20 trials: 10 for ascending and 10 for descending. The quantitative outcomes are summarized in TABLE \ref{hardwareResultsStairs}. Comparatively, ascending stairs proved more challenging, requiring a longer average duration (22.65 s) and incurring higher risk (1 fall). 

\section{Discussions} \label{discussions}
While our extensive simulation benchmarks and ablation studies demonstrate the framework's effectiveness, several sim-to-real discrepancies remain. Stair environments proved particularly challenging, as the simultaneous demand for balancing and dynamic obstacle avoidance pushed the policy to its limits. Furthermore, despite fusing IMU data, we observed that high-frequency vibrations from quadrupedal locomotion often induced localization drift in the LiDAR SLAM algorithm, degrading real-world navigation accuracy. Integrating an uncertainty-aware frame alignment mechanism—such as weighting historical scans by pose covariance—or adopting direct relative-motion visual/LiDAR odometry presents a promising avenue to enhance policy robustness against severe localization drift on uneven terrain. Another critical hardware constraint is the limited Field of View (FoV) of the onboard camera; unlike the $360^\circ$ FoV available in simulation, our hardware deployment assumes pedestrians outside the camera's range will actively avoid the robot. Addressing these sensing and environmental gaps is a primary objective for future development.

The modeling of human behavior presents an additional limitation. While we utilize ORCA to simulate multi-agent interactions, which is consistent with established literature, this model does not fully capture the complexity and diversity of real-world human interactions. Such discrepancies can impede the transferability of the navigation policy from simulated environments to real-world spaces. Consequently, future research will focus on integrating more sophisticated and heterogeneous interaction models into our simulation pipeline to enhance the robustness and reliability during practical deployment. Furthermore, formulating constrained bi-level optimization for joint locomotion and navigation policies presents a compelling avenue for future research.

\section{Conclusions} \label{conclusions}
In this work, we presented a unified locomotion-navigation policy with interaction awareness for quadruped robots operating on uneven terrains. Our framework employs a two-stage training strategy: first, we independently developed a high-level motion planning policy for interactive environments using a differential-drive kinematic model and a low-level locomotion controller tailored to specific quadrupedal dynamics. In the second stage, these components were integrated into a single, interaction-aware policy capable of simultaneously managing complex pedestrian interactions and uneven terrain traversal.

Comprehensive benchmarks and ablation studies validated the effectiveness of the proposed architecture, while extensive real-world deployments demonstrated its robust sim-to-real transfer potential across diverse environments. To address current limitations, future work will focus on enhancing locomotion stability, specifically for challenging tasks like stair traversal. Additionally, we aim to improve the realism of our simulation environments by integrating diverse and heterogeneous pedestrian interaction models to broadly reflect the complexity of real-world human behavior.

\appendix
\subsection{Locomotion Networks} \label{locomotionNetworkDetail}
TABLEs \ref{locomotionActorCriticDecoderNetworks} and \ref{locomotionEncoderNetworks} detail the locomotion networks.

\renewcommand\arraystretch{1.2}
\begin{table}[!t]
	\small
	\centering
	\caption{Locomotion networks: actor, critic, and decoder}
	\begin{tabular}{c c c c} 
		\hline
		Networks  & Input shape   & Hidden layers     & Output shape \\
		\hline 
		Actor     & (64,)         & $512 \times 256 \times 128$     & (12,) \\
            Critic    & (286,)        & $512 \times 256 \times 128$     & (1,)\\
            Decoder   & (19,)         & $64 \times 128$                 & (45,)\\
		\hline
	\end{tabular}
	\label{locomotionActorCriticDecoderNetworks}
\end{table}

\renewcommand\arraystretch{1.2}
\begin{table}[!t]
	\small
	\centering
	\caption{Locomotion networks: encoder}
	\begin{tabular}{c c} 
		\hline
		Layer     & Operation   \\
		\hline 
		1         & Conv2d(1, 16, kernel$_-$size=3, stride=1, padding=1)     \\
            2         & MaxPool2d(kernel$_-$size=2, stride=2, padding=0)    \\
            3         & Conv2d(16, 32, kernel$_-$size=3, stride=1, padding=1)     \\
            4         & MaxPool2d(kernel$_-$size=2, stride=2, padding=0)    \\
            5         & Conv2d(32, 64, kernel$_-$size=3, stride=1, padding=1)     \\
            6         & Flatten with output shape (1408,)  \\
            7         & Linear(1408, 64)  \\
            Output    & 2 $\times$ Linear(64, 3), 2 $\times$ Linear(64, 16)\\
		\hline
	\end{tabular}
	\label{locomotionEncoderNetworks}
\end{table}

\subsection{Navigation Networks} \label{navigationNetworkDetail}
TABLEs \ref{navigationActorCriticNetworks} and \ref{navigationEncoderNetworks} detail the navigation networks.

\renewcommand\arraystretch{1.2}
\begin{table}[!t]
	\small
	\centering
	\caption{Navigation networks: actor, critic, and decoder}
	\begin{tabular}{c c c c} 
		\hline
		Networks  & Input shape   & Hidden layers         & Output shape \\
		\hline 
		Actor     & (54)          & $1024 \times 1024$     & (4,) \\
            Critic    & (56,)         & $1024 \times 1024$     & (1,)\\
		\hline
	\end{tabular}
	\label{navigationActorCriticNetworks}
\end{table}

\renewcommand\arraystretch{1.2}
\begin{table}[!t]
	\small
	\centering
	\caption{Navigation networks: encoder}
	\begin{tabular}{c c} 
		\hline
		Layer     & Operation   \\
		\hline 
		1         & Conv2d(1, 32, kernel$_-$size=(2, 20), stride=(1, 10))     \\
            2         & MaxPool2d(kernel$_-$size=(1, 5))    \\
            3         & Conv2d(32, 32, kernel$_-$size=2, stride=1)     \\
            4         & MaxPool2d(kernel$_-$size=(1, 2))    \\
            5         & Flatten with output shape (2176,)  \\
            6         & Linear(2176, 1024)  \\
            Output    & (50,)\\
		\hline
	\end{tabular}
	\label{navigationEncoderNetworks}
\end{table}

\bibliography{references}

\end{document}